\documentclass[sigconf]{acmart}

\copyrightyear{2020}
\acmYear{2020}
\setcopyright{acmcopyright}
\acmConference[ICSE '20]{42nd International Conference on Software Engineering}{May 23--29, 2020}{Seoul, Republic of Korea}
\acmBooktitle{42nd International Conference on Software Engineering (ICSE '20), May 23--29, 2020, Seoul, Republic of Korea}
\acmPrice{15.00}
\acmDOI{10.1145/3377811.3380337}
\acmISBN{978-1-4503-7121-6/20/05}

\startPage{1}

\usepackage{tikz}

\usepackage{algorithm2e}
\usepackage{amsmath}

\newcommand{\pluseq}{\mathrel{+}=}

\definecolor{mygreen}{rgb}{0,0.6,0}
\definecolor{mygray}{rgb}{0.5,0.5,0.5}
\definecolor{mymauve}{rgb}{0.58,0,0.82}
\definecolor{dkgreen}{rgb}{0,0.5,0}

\definecolor{lightgray}{rgb}{0.85,0.85,0.85}
\definecolor{lightgreen}{rgb}{0.7,0.9,0.7}
\definecolor{lightblue}{rgb}{0.7,0.7,0.9}
\definecolor{lightred}{rgb}{0.9,0.7,0.7}

\newcommand{\ignore}[1]{}

\newcommand{\proc}[1]{\textsc{#1}}

\newcommand{\diffNN}{\textsc{ReluDiff}}
\newcommand{\ReluVal}{\textsc{ReluVal}}
\newcommand{\DeepPoly}{\textsc{DeepPoly}}

\newcommand{\Reluplex}{\textsc{Reluplex}}

\newcommand{\RefineZono}{\textsc{RefineZono}}

\newcommand\UB[1]{\mathsf{UB}\big(#1\big)}
\newcommand\LB[1]{\mathsf{LB}\big(#1\big)}

\newcommand\UBconcrete[1]{\overline{\mathsf{UB}}\big(#1\big)}
\newcommand\LBconcrete[1]{\underline{\mathsf{LB}}\big(#1\big)}

\newcommand\ReLU[1]{\proc{ReLU}\left(#1\right)}
\newcommand\ReluOUT[1]{S(#1)}
\newcommand\ReluIN[1]{S^{in}(#1)}
\newcommand\ReluOUTdelta[1]{\delta(#1)}
\newcommand\ReluINdelta[1]{\delta^{in}(#1)}
\newcommand\EdgeOUTdelta[1]{\delta(#1)}

\newcommand\n[0]{n_{k,j}}
\newcommand\np[0]{n'_{k,j}}
\newcommand{\deq}{\Delta}

\begin{document}

\title[]{ReluDiff: Differential Verification of Deep Neural Networks}

\author{Brandon Paulsen}
\affiliation{
  \institution{University of Southern California}
  \city{Los Angeles}
  \state{California}
  \country{USA} 
}
\author{Jingbo Wang}
\affiliation{
  \institution{University of Southern California}
  \city{Los Angeles}
  \state{California}
  \country{USA} 
}
\author{Chao Wang}
\affiliation{
  \institution{University of Southern California}
  \city{Los Angeles}
  \state{California}
  \country{USA}     
}

\begin{abstract}
As deep neural networks are increasingly being deployed in practice,
their efficiency has become an important issue.  While there are
compression techniques for reducing the network's size, energy
consumption and computational requirement, they only
demonstrate \textit{empirically} that there is no loss of accuracy,
but lack formal guarantees of the compressed network, e.g., in the
presence of adversarial examples.
Existing verification techniques such as \Reluplex{}, \ReluVal{}, and
\DeepPoly{} provide formal guarantees, but they are designed for
analyzing a single network instead of the relationship between two
networks.
To fill the gap, we develop a new method for \emph{differential
verification} of two closely related networks.
Our method consists of a fast but approximate \emph{forward interval
analysis pass} followed by a \emph{backward pass} that iteratively refines the
approximation until the desired property is verified.  We have two main innovations.  During the forward
pass, we exploit structural and behavioral similarities of the two
networks to more accurately bound the difference between the output
neurons of the two networks. Then in the backward pass, we leverage 
the gradient differences to more accurately compute the most beneficial
refinement.
Our experiments show that, compared to state-of-the-art verification
tools, our method can achieve orders-of-magnitude speedup and prove
many more properties than existing tools.
\end{abstract}

\maketitle

\section{Introduction}

As deep neural networks (DNNs) make their way into safety critical
systems such as aircraft collision avoidance~\cite{JulianKO18} and
autonomous driving \cite{bojarski2016end}, where errors may lead to
catastrophes, there is a growing need for formal verification.  The
situation is further exacerbated by \textit{adversarial
examples}~\cite{szegedy2013intriguing,GoodfellowSS15}, which are
security exploits created specifically to cause erroneous
classifications~\cite{NguyenYC15, XuQE16, Moosavi-Dezfooli16,
KurakinGB17a}.
There is also a growing need for reducing the size of the neural
networks deployed on energy- and computation-constrained devices.
Consequently, compression techniques~\cite{HanMD16} have emerged to
prune unnecessary edges, quantize the weights of remaining edges, and
retrain the networks, but they do not provide any formal guarantee --
typically the accuracy of a compressed network is only
demonstrated \textit{empirically}.

While empirical evidence or statistical analysis may increase our
confidence that a network behaves as expected for most of the inputs,
they cannot prove that it does so for all inputs.
Similarly, while heuristic search and dynamic analysis techniques,
including testing~\cite{PeiCYJ17, TianPJR18, ma2018deepgauge} and
fuzzing~\cite{odena2018tensorfuzz, xie2019deephunter, xie2019diffchaser}, may quickly discover adversarial
examples, they cannot prove the absence of such examples.
At the same time, while state-of-the-art verification
techniques~\cite{HuangKWW17,Ehlers17,KatzHIJLLSTWZDK19,RuanHK18,WangPWYJ18nips,SinghGPV19iclr,MirmanGV18,GehrMDTCV18,FischerBDGZV19},
including \Reluplex{}~\cite{KatzBDJK17}, \ReluVal{}~\cite{WangPWYJ18}
and \DeepPoly{}~\cite{SinghGPV19}, can provide formal proofs, they are
designed for analyzing a single network as opposed to the relationship
between two networks.

\begin{figure}
\centering
\scalebox{0.9}{\input{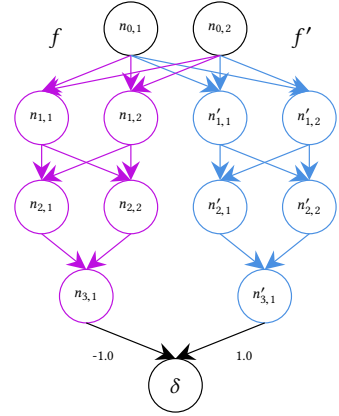}}
\caption{Differential verification of deep neural networks.} 
\label{fig:subbed_nnets}
\end{figure}

In this work, we focus on \textit{differential verification} of two
closely related networks.  In this problem domain, we assume that
$ f $ and $ f' $ are two neural networks trained for the same task;
that is, they accept the same input $x$ and are expected to produce
the same output.  They are also structurally the same while differing
only in the numerical values of edge weights (which allows us to 
analyze compression techniques such as quantization and edge pruning~\cite{HanMD16}).  
In this context, differential verification is concerned with proving
 $\forall x \in X $. $ |f'(x) - f(x)| < \epsilon $, where $ X $
is an input region of interest and $ \epsilon $ is some reasonably
small bound.  
This problem has not received adequate attention and, as we will show
in this work, existing tools are ill-suited for solving this problem.

The key limitation of existing tools is that, since they are designed
to analyze the behavior of a single network, they do not have the
ability to exploit the structural similarities of two closely related
networks.  They also have difficulty handling the constraint that the inputs
to both $ f $ and $ f' $ are identical.
Typically, these tools work by computing the conservative value ranges
of all neurons from input to output in a layer-by-layer style. In the
early layers, they may be able to maintain relationships between the
inputs, but as the functions become increasingly non-linear in
subsequent layers, approximations must be made.  This ``eager''
approximation means relationships between the inputs of $ f $ and $ f'
$ are mostly lost, causing extremely large over-approximations in
the output layer.

In fact, state-of-the-art verification tools that we have investigated
(\ReluVal{}~\cite{WangPWYJ18} and \DeepPoly{}~\cite{SinghGPV19})
struggle to verify that \emph{two identical networks are the same}.
To carry out this \emph{litmus test} without drastically altering
these tools, we construct a combined network as shown in
Figure~\ref{fig:subbed_nnets}, where $f$ and $f'$ are actually the
same network (i.e. same structure and edge weights).  Since they share the same input
$x$, we expect $f(x)-f(x)$ to be 0 regardless of the input region for
$x$.  While our method can easily prove that $|f(x)-f(x)|< \epsilon$
for an arbitrarily small $\epsilon$ in less than a second,
none of the existing tools are able to do so.  In fact, \DeepPoly{}
cannot verify it no matter how much time is given (it is not a
complete method) and
\ReluVal{} times out after several hours.

\begin{figure}[t]
\scalebox{0.75}{\input{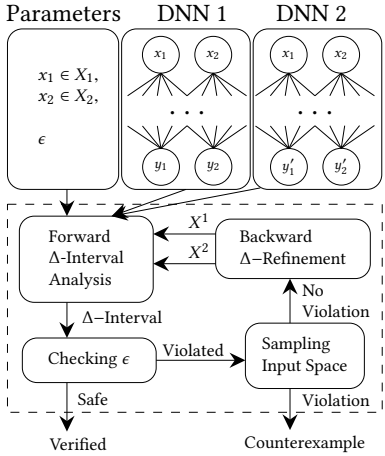}} 
\caption{The differential verification flow of \diffNN{}.} 
\label{fig:diagram}
\end{figure}

Figure~\ref{fig:diagram} shows the overall flow of our method,
\diffNN{}, whose  input consists of the two networks (DNN1 and DNN2),
an input region ($x_1\in X_1$ and $x_2\in X_2$), and a bound
$\epsilon$ on the output difference.  
There are three possible outcomes: (1) \emph{verified}, meaning that
the output difference is proved to be less than $\epsilon$;
(2) \emph{falsified}, meaning a counterexample is found; or
(3) \emph{unknown}, meaning that verification remains inconclusive due
to bounds on the computing resources.

Internally, \diffNN{} iterates through two steps: a forward pass and a
backward pass.
The forward pass computes over-approximated value differences of
corresponding neurons in the two networks, and propagates them layer
by layer from the input to the output.
If the output difference is within the region $[-\epsilon, \epsilon]
$, the property is verified.
Otherwise, \diffNN{} samples a fixed number of concrete examples from
the input space and tests if they violate the property. If a violation
is found, the property is falsified; otherwise, \diffNN{} enters the
refinement phase.

The goal of refinement is to identify an input region that should be
divided into subregions.  By using these subregions to perform the
forward pass again, some of the forced over-approximations may be
avoided, thus leading to significant accuracy increase.
To identify the right input region for refinement, the \emph{backward
pass} computes the difference of the gradients of the two networks and
uses it to find input regions that, once divided into subregions, are
more likely to result in accuracy increase.

While iterative interval analysis has been used in verifying neural
networks before~\cite{WangPWYJ18}, the focus has always been on a single network.  In
this work, we show that, by focusing on both networks simultaneously,
we can be more efficient and accurate compared to analyzing each
network in isolation.
Note that, in differential verification, the two networks have
identical structures and similar behaviors; therefore, we can easily
develop a correspondence between neurons in $ f $ and $ f' $, thus
allowing a lock-step style verification.  Lock-step verification
allows us to directly compute the differences in values of neurons and
propagate these differences through edges. It also allows symbolic
intervals to be used to avoid some of the approximations.
Since error caused by approximation grows quickly, sometimes
exponentially~\cite{WangPWYJ18nips}, as it is propagated through edges
and neurons, this can significantly increase accuracy.

When approximation must be made, e.g., due to non-linearity of ReLU,
we can handle them better by focusing on the value differences instead
of the absolute values.  For example, in \ReluVal{}~\cite{WangPWYJ18},
if a symbolic expression that represents the ReLU input may be both
positive and negative, the symbolic expression must be replaced by an
interval with concrete upper and lower bounds, which introduces
additional error.  In contrast, we can be more accurate: even if the
input value of a neuron may be both positive and negative, in many
cases we still can avoid introducing error into the difference.

We have implemented \diffNN{} in a tool and evaluated it on a number
of feed-forward neural network benchmarks, including ACAS Xu for
aircraft collision avoidance~\cite{JulianKO18}, MNIST for hand-written
digit recognition~\cite{LecunBBH98}, and HAR for human activity
recognition~\cite{AnguitaHAR}.  We also experimentally
compared \diffNN{} with state-of-the-art tools,
including \ReluVal{}~\cite{WangPWYJ18}
and \DeepPoly{}~\cite{SinghGPV19}.  Our experimental results show
that, in almost all cases, \diffNN{} outperforms these existing tools
in both speed and accuracy.
In total, we evaluate on 842 properties over our benchmark
networks. \diffNN{} was often one to two orders-of-magnitude faster,
and was able to prove 745 out of the 842 properties whereas none of
the other tools can prove more than 413 properties.

To summarize, we make the following contributions:
\begin{itemize}
\item 
We propose the first iterative symbolic interval analysis for
differential verification of two neural networks.
\item 
We develop a forward pass algorithm for more accurately computing the
value differences for corresponding neurons.
\item
We develop a backward pass algorithm, based on gradient difference,
for computing the refinement.
\item 
We implement the method and demonstrate its advantages over existing
tools in terms of both speed and accuracy.
\end{itemize}

The remainder of the paper is organized as follows.  First, we use
examples to motivate our method in Section~\ref{sec:motivation}.
Then, we review the basics of neural networks and interval analysis in
Section~\ref{sec:prelims}.  Next, we present our method for the
forward pass in Section~\ref{sec:forward-pass}, followed by our method
for the backward pass in Section~\ref{sec:backward-pass}.  We present
our experimental results in Section~\ref{sec:experiment}.  We review
the related work in Section~\ref{sec:related}.  Finally, we give our
conclusions in Section~\ref{sec:conclusion}.

\section{Motivation}
\label{sec:motivation}

We illustrate the problems of existing verification tools using
examples and then highlight our main contributions.

\subsection{Differential Verification}

\begin{figure}
\scalebox{0.65}{\tikzset{every picture/.style={line width=0.75pt}} %set default line width to 0.75pt        

\begin{tikzpicture}[x=0.75pt,y=0.75pt,yscale=-1,xscale=1]
%uncomment if require: \path (0,300); %set diagram left start at 0, and has height of 300

%Shape: Circle [id:dp29452950296941915] 
\draw   (90,55) .. controls (90,41.19) and (101.19,30) .. (115,30) .. controls (128.81,30) and (140,41.19) .. (140,55) .. controls (140,68.81) and (128.81,80) .. (115,80) .. controls (101.19,80) and (90,68.81) .. (90,55) -- cycle ;
%Shape: Circle [id:dp7730288688046164] 
\draw   (90,160) .. controls (90,146.19) and (101.19,135) .. (115,135) .. controls (128.81,135) and (140,146.19) .. (140,160) .. controls (140,173.81) and (128.81,185) .. (115,185) .. controls (101.19,185) and (90,173.81) .. (90,160) -- cycle ;
%Shape: Circle [id:dp7848793538270885] 
\draw   (190,160) .. controls (190,146.19) and (201.19,135) .. (215,135) .. controls (228.81,135) and (240,146.19) .. (240,160) .. controls (240,173.81) and (228.81,185) .. (215,185) .. controls (201.19,185) and (190,173.81) .. (190,160) -- cycle ;
%Shape: Circle [id:dp2581761932666119] 
\draw   (190,55) .. controls (190,41.19) and (201.19,30) .. (215,30) .. controls (228.81,30) and (240,41.19) .. (240,55) .. controls (240,68.81) and (228.81,80) .. (215,80) .. controls (201.19,80) and (190,68.81) .. (190,55) -- cycle ;
%Shape: Circle [id:dp03162623697960232] 
\draw   (290,160) .. controls (290,146.19) and (301.19,135) .. (315,135) .. controls (328.81,135) and (340,146.19) .. (340,160) .. controls (340,173.81) and (328.81,185) .. (315,185) .. controls (301.19,185) and (290,173.81) .. (290,160) -- cycle ;
%Shape: Circle [id:dp02546394024001375] 
\draw   (290,55) .. controls (290,41.19) and (301.19,30) .. (315,30) .. controls (328.81,30) and (340,41.19) .. (340,55) .. controls (340,68.81) and (328.81,80) .. (315,80) .. controls (301.19,80) and (290,68.81) .. (290,55) -- cycle ;
%Shape: Circle [id:dp12084775666404268] 
\draw   (371,105) .. controls (371,91.19) and (382.19,80) .. (396,80) .. controls (409.81,80) and (421,91.19) .. (421,105) .. controls (421,118.81) and (409.81,130) .. (396,130) .. controls (382.19,130) and (371,118.81) .. (371,105) -- cycle ;
%Straight Lines [id:da34998381745971496] 
\draw    (140,160) -- (188,160) ;
\draw [shift={(190,160)}, rotate = 180] [fill={rgb, 255:red, 0; green, 0; blue, 0 }  ][line width=0.75]  [draw opacity=0] (10.72,-5.15) -- (0,0) -- (10.72,5.15) -- (7.12,0) -- cycle    ;

%Straight Lines [id:da4638805061238097] 
\draw    (140,160) -- (189.14,56.81) ;
\draw [shift={(190,55)}, rotate = 475.46] [fill={rgb, 255:red, 0; green, 0; blue, 0 }  ][line width=0.75]  [draw opacity=0] (10.72,-5.15) -- (0,0) -- (10.72,5.15) -- (7.12,0) -- cycle    ;

%Straight Lines [id:da15742477491714235] 
\draw    (140,55) -- (188,55) ;
\draw [shift={(190,55)}, rotate = 180] [fill={rgb, 255:red, 0; green, 0; blue, 0 }  ][line width=0.75]  [draw opacity=0] (10.72,-5.15) -- (0,0) -- (10.72,5.15) -- (7.12,0) -- cycle    ;

%Straight Lines [id:da9054930639679811] 
\draw    (140,55) -- (189.14,158.19) ;
\draw [shift={(190,160)}, rotate = 244.54000000000002] [fill={rgb, 255:red, 0; green, 0; blue, 0 }  ][line width=0.75]  [draw opacity=0] (10.72,-5.15) -- (0,0) -- (10.72,5.15) -- (7.12,0) -- cycle    ;

%Straight Lines [id:da0727882271401632] 
\draw    (240,160) -- (289.14,56.81) ;
\draw [shift={(290,55)}, rotate = 475.46] [fill={rgb, 255:red, 0; green, 0; blue, 0 }  ][line width=0.75]  [draw opacity=0] (10.72,-5.15) -- (0,0) -- (10.72,5.15) -- (7.12,0) -- cycle    ;

%Straight Lines [id:da02068180741540726] 
\draw    (240,55) -- (289.14,158.19) ;
\draw [shift={(290,160)}, rotate = 244.54000000000002] [fill={rgb, 255:red, 0; green, 0; blue, 0 }  ][line width=0.75]  [draw opacity=0] (10.72,-5.15) -- (0,0) -- (10.72,5.15) -- (7.12,0) -- cycle    ;

%Straight Lines [id:da4941096166166473] 
\draw    (340,55) -- (369.95,103.3) ;
\draw [shift={(371,105)}, rotate = 238.2] [fill={rgb, 255:red, 0; green, 0; blue, 0 }  ][line width=0.75]  [draw opacity=0] (10.72,-5.15) -- (0,0) -- (10.72,5.15) -- (7.12,0) -- cycle    ;

%Straight Lines [id:da0005753242144755921] 
\draw    (340,160) -- (370.02,106.74) ;
\draw [shift={(371,105)}, rotate = 479.41] [fill={rgb, 255:red, 0; green, 0; blue, 0 }  ][line width=0.75]  [draw opacity=0] (10.72,-5.15) -- (0,0) -- (10.72,5.15) -- (7.12,0) -- cycle    ;

%Straight Lines [id:da6663993224468625] 
\draw    (240,160) -- (288,160) ;
\draw [shift={(290,160)}, rotate = 180] [fill={rgb, 255:red, 0; green, 0; blue, 0 }  ][line width=0.75]  [draw opacity=0] (10.72,-5.15) -- (0,0) -- (10.72,5.15) -- (7.12,0) -- cycle    ;

%Straight Lines [id:da971612493075633] 
\draw    (240,55) -- (288,55) ;
\draw [shift={(290,55)}, rotate = 180] [fill={rgb, 255:red, 0; green, 0; blue, 0 }  ][line width=0.75]  [draw opacity=0] (10.72,-5.15) -- (0,0) -- (10.72,5.15) -- (7.12,0) -- cycle    ;

% Text Node
\draw (162.5,39) node [scale=1] [align=left] {1.9};
% Text Node
\draw (132.5,90) node [scale=1] [align=left] {1.1};
% Text Node
\draw (130,120) node [scale=1] [align=left] {\mbox{-}2.1};
% Text Node
\draw (161.5,171) node [scale=1] [align=left] {1.0};
% Text Node
\draw (264.5,39) node [scale=1] [align=left] {2.1};
% Text Node
\draw (238,90) node [scale=1] [align=left] {\mbox{-}0.9};
% Text Node
\draw (252.5,171) node [scale=1] [align=left] {1.1};
% Text Node
\draw (238,120) node [scale=1] [align=left] {\mbox{-}1.0};
% Text Node
\draw (372.5,69) node [scale=1] [align=left] {1.0};
% Text Node
\draw (375,141) node [scale=1] [align=left] {\mbox{-}1.0};
% Text Node
\draw (115,55) node   {$n_{0,1}$};
% Text Node
\draw (115,160) node  {$n_{0,2}$};
% Text Node
\draw (215,160) node   {$n_{1,2}$};
% Text Node
\draw (215,55) node   {$n_{1,1}$};
% Text Node
\draw (315,55) node   {$n_{2,1}$};
% Text Node
\draw (315,160) node   {$n_{2,2}$};
% Text Node
\draw (396,105) node   {$n_{3,1}$};

% Text Node
\draw (70,55) node   {$x_1$};
% Text Node
\draw (70,160) node  {$x_2$};
% Text Node
\draw (470,105) node   {$y = f (x_1,x_2)$};

\end{tikzpicture}} 
\caption{A neural network with two inputs and one output.}
\label{fig:single-network}
\end{figure}
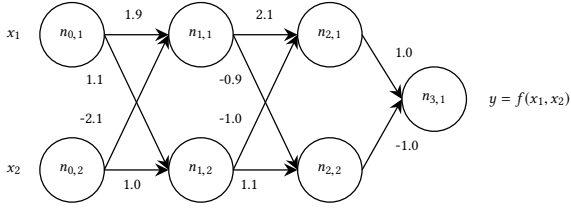

Figure~\ref{fig:single-network} shows a feed-forward neural network
with one input layer, two hidden layers, and one output layer.  
The input layer has two nodes $n_{0,1}$ and $n_{0,2}$, corresponding to
the two input variables $x_1$ and $x_2$.
Each hidden layer consists of two neurons, $n_{1,1}, n_{1,2}$ in one
layer and $n_{2,1},n_{2,2}$ in the other layer.  Each of these neurons
has two computation steps: the affine transformation and the ReLU
activation.  For example, inside $n_{1,1}$, the affine transformation
is $1.9 x_1 -2.1 x_2$ and the ReLU activation is $max(0, 1.9 x_1 - 2.1
x_2)$.
The output layer has one node, representing the value of $y =
f(x_1,x_2)$. In general, $f$ is a non-linear function over $x_1$ and
$x_2$.

In differential verification, we are concerned with the relationship
between $f(x_1,x_2)$ and another network $f'(x_1,x_2)$.  For the sake
of example, we focus on a network compression technique
called \textit{quantiziation}~\cite{HanMD16} in which the edge weights
of $f$ are rounded to the nearest whole number to obtain $f'$.
However, we note that our method can be used on \textit{any} two
networks with similar structures, e.g., when $f'$ is created using
other techniques including \emph{edge pruning} and \emph{network
retraining}~\cite{HanMD16,HeLLWLH18,JulianKO18,SehwagWMJ19}.

These techniques, in general, raise the concern on how they affect the
network's behavior.  In particular, we would like to verify that the
new network produces outputs within some bound relative to the
original network.  Formally, let $ f' : \mathbb{X} \to \mathbb{Y} $ be
the second network and $ f : \mathbb{X} \to \mathbb{Y} $ be the first
network. We would like to verify that $ |f'(x) - f(x)| < \epsilon $
for all $ x \in X $, where $ X \subseteq \mathbb{X}$ is some region of
importance in the input domain $\mathbb{X}$.

\subsection{Existing Approaches}

Existing tools for verifying neural networks target only a single
network at a time, and are often geared toward proving the absence
of \textit{adversarial examples}.
That is, given an input region of interest, they decide if the output
stays in a desired region.  For the network in
Figure~\ref{fig:single-network}, in particular, the input region may
be $x_1\in[4,6]$ and $x_2\in [1,5]$, and the desired output
may be $f(x_1,x_2) < 15$.
However, these tools are not designed for verifying the relationship
between two networks.  While we could try and re-use them for our
purpose, they lack the ability to exploit the similarities of the two
networks.

For example, we could use the existing
tool \ReluVal{}~\cite{WangPWYJ18} on both $ f $ and $ f' $ to compare
the concrete output intervals it computes for an input region of
interest, e.g., $ [f_{low}, f_{up}] $ and $ [f'_{low}, f'_{up}]$.  In
order to conservatively estimate the difference between $ f $ and $ f'
$, we must assume the maximum difference falls in the interval $
[f'_{low} - f_{up}, f'_{up} - f_{low}] $.
In Figure~\ref{fig:single-network}, the interval difference would be
$[-25.76, 22.93]$, which is too large to be useful.

Even though \ReluVal{} could tighten the interval by refining the
input intervals, this naive approach cannot even verify that two
identical networks always produce the same output, since the output
intervals do not capture that the corresponding inputs to $ f $ and $
f' $ (i.e., values of $x_1$ and $x_2$) are always the same.

\begin{figure}
\centering
\scalebox{0.65}{\input{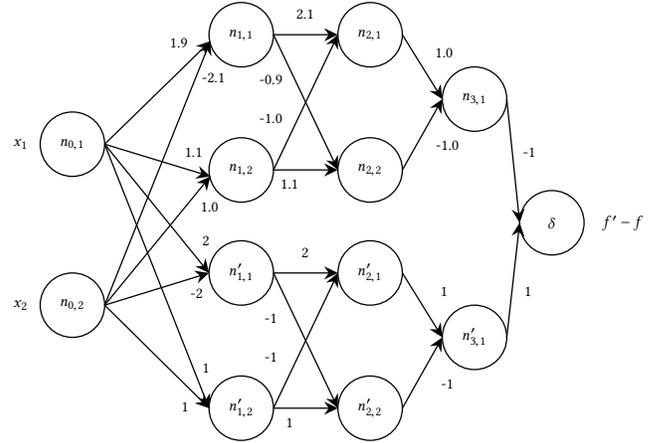}}
\caption{Naive differential verification of the two networks.}
\label{fig:composed-network}
\end{figure}

To compensate, we could encode the constraint that values of the
corresponding inputs are always the same by composing $ f $ and $ f' $
into a single feed-forward network equivalent to $ f' - f $, as shown
in Figure~\ref{fig:composed-network}. In theory, a sound and complete
technique would be able to verify, \textit{eventually}, that the
output difference is bounded by an arbitrarily small $\epsilon$, but
with a caveat.

That is, to maintain the relationships between the input variables and
the difference in the outputs of the two networks, each neuron must
remain in a linear state across the entire input region; otherwise,
approximation must be made to maintain the soundness of the interval
analysis.  However, approximation inevitably loses some of the
relationships between the inputs and the outputs.
Indeed, we constructed some merged 
networks in the same way as in Figure~\ref{fig:composed-network} and
then fed them to existing tools.  Unfortunately, they all exhibit the
``worst-case'' value range blowup in the output.

%To be concrete, consider our attempt to verify that $ f(x_1,x_2) -
%f'(x_1,x_2) < \epsilon $ using \ReluVal{}~\cite{WangPWYJ18}, which
%uses symbolic intervals to represent the value ranges whenever
%possible.  That is, upper and lower bounds of these intervals are
%represented as linear functions of $x_1$ and $x_2$.  However, due to
%the non-linear nature of $f$ and $f'$, approximations have to be made
%during the forward analysis, which result in the maximum absolute
%difference being 21.36.  This is far more than the actual maximum of
%$ \approx 4.945 $. 

The key reason is that existing tools such as \ReluVal{} are forced to
approximate ReLU activations by concretizing, which is then followed by 
interval subtractions, thus causing error introduced by these approximations
to be quickly amplified.  
The forward pass over $ f $ computes an output interval of $ [-1.2 x_1
- 1.1 x_2, -1.1 x_1 - x_2 + 19.53] $, and for $ f' $ it computes $
[-x_1 - x_2, -x_1 - x_2 + 20] $.  Although the equations are symbolic,
the difference $[-21.36, 19.13]$, computed conservatively
by \ReluVal{}, is still too large to be useful.

\subsection{Our Method}

Existing tools cannot exploit structural and behavioral similarities
of the two networks in differential verification.  Our insight is to
leverage such similarities to drastically improve both the efficiency
and the accuracy of the verification tool.

Specifically, in this work, we pair neurons and edges of the first
network with those of the second network and then perform
a \emph{lock-step} verification.  This allows us to focus on the value
differences of the corresponding neurons as opposed to their absolute
values. The benefit is that doing so results in both fewer and tighter approximations
and more error reduction due to the use of symbolic intervals.
We also perform better refinement by focusing on inputs that have the
greatest influence on the output difference, rather than the
absolute output values.

While focusing on the \emph{difference} as opposed to \emph{absolute
values} \textit{seems} to be a straightforward idea, there are many technical challenges.
For example, there will be significantly more complex ReLU activation
patterns to consider since we have to handle both networks
simultaneously, instead of one network at a time.  Approximating
symbolic intervals when considering the output difference of two ReLU
activations (i.e., ReLU($ x' $) - ReLU($ x $)) has yet to be studied
and is non-trivial.  Furthermore, how to determine which input neuron
to refine when the goal is to reduce error in the output difference
between two networks has not been considered either.

In this work, we develop solutions to overcome these challenges.
During forward interval analysis, we carefully consider the ReLU
activation patterns, and propose a technique for handling each pattern
soundly while minimizing the approximation error.  During the
refinement, we compute the difference between gradients of the two
networks, and use it to identify the input neuron most likely to
increase the accuracy of the differential verification result.

\begin{figure}
\centering
\scalebox{0.7}{\tikzset{every picture/.style={line width=0.75pt}} %set default line width to 0.75pt        

\begin{tikzpicture}[x=0.75pt,y=0.75pt,yscale=-1,xscale=1]
%uncomment if require: \path (0,300); %set diagram left start at 0, and has height of 300

%Shape: Circle [id:dp29452950296941915] 
\draw   (90,96) .. controls (90,82.19) and (101.19,71) .. (115,71) .. controls (128.81,71) and (140,82.19) .. (140,96) .. controls (140,109.81) and (128.81,121) .. (115,121) .. controls (101.19,121) and (90,109.81) .. (90,96) -- cycle ;
%Shape: Circle [id:dp7730288688046164] 
\draw   (90,201) .. controls (90,187.19) and (101.19,176) .. (115,176) .. controls (128.81,176) and (140,187.19) .. (140,201) .. controls (140,214.81) and (128.81,226) .. (115,226) .. controls (101.19,226) and (90,214.81) .. (90,201) -- cycle ;
%Shape: Circle [id:dp7848793538270885] 
\draw   (190,201) .. controls (190,187.19) and (201.19,176) .. (215,176) .. controls (228.81,176) and (240,187.19) .. (240,201) .. controls (240,214.81) and (228.81,226) .. (215,226) .. controls (201.19,226) and (190,214.81) .. (190,201) -- cycle ;
%Shape: Circle [id:dp2581761932666119] 
\draw   (190,96) .. controls (190,82.19) and (201.19,71) .. (215,71) .. controls (228.81,71) and (240,82.19) .. (240,96) .. controls (240,109.81) and (228.81,121) .. (215,121) .. controls (201.19,121) and (190,109.81) .. (190,96) -- cycle ;
%Shape: Circle [id:dp03162623697960232] 
\draw   (290,201) .. controls (290,187.19) and (301.19,176) .. (315,176) .. controls (328.81,176) and (340,187.19) .. (340,201) .. controls (340,214.81) and (328.81,226) .. (315,226) .. controls (301.19,226) and (290,214.81) .. (290,201) -- cycle ;
%Shape: Circle [id:dp02546394024001375] 
\draw   (290,96) .. controls (290,82.19) and (301.19,71) .. (315,71) .. controls (328.81,71) and (340,82.19) .. (340,96) .. controls (340,109.81) and (328.81,121) .. (315,121) .. controls (301.19,121) and (290,109.81) .. (290,96) -- cycle ;
%Shape: Circle [id:dp12084775666404268] 
\draw   (371,146) .. controls (371,132.19) and (382.19,121) .. (396,121) .. controls (409.81,121) and (421,132.19) .. (421,146) .. controls (421,159.81) and (409.81,171) .. (396,171) .. controls (382.19,171) and (371,159.81) .. (371,146) -- cycle ;
%Straight Lines [id:da34998381745971496] 
\draw    (140,201) -- (188,201) ;
\draw [shift={(190,201)}, rotate = 180] [fill={rgb, 255:red, 0; green, 0; blue, 0 }  ][line width=0.75]  [draw opacity=0] (10.72,-5.15) -- (0,0) -- (10.72,5.15) -- (7.12,0) -- cycle    ;

%Straight Lines [id:da4638805061238097] 
\draw    (140,201) -- (189.14,97.81) ;
\draw [shift={(190,96)}, rotate = 475.46] [fill={rgb, 255:red, 0; green, 0; blue, 0 }  ][line width=0.75]  [draw opacity=0] (10.72,-5.15) -- (0,0) -- (10.72,5.15) -- (7.12,0) -- cycle    ;

%Straight Lines [id:da15742477491714235] 
\draw    (140,96) -- (188,96) ;
\draw [shift={(190,96)}, rotate = 180] [fill={rgb, 255:red, 0; green, 0; blue, 0 }  ][line width=0.75]  [draw opacity=0] (10.72,-5.15) -- (0,0) -- (10.72,5.15) -- (7.12,0) -- cycle    ;

%Straight Lines [id:da9054930639679811] 
\draw    (140,96) -- (189.14,199.19) ;
\draw [shift={(190,201)}, rotate = 244.54000000000002] [fill={rgb, 255:red, 0; green, 0; blue, 0 }  ][line width=0.75]  [draw opacity=0] (10.72,-5.15) -- (0,0) -- (10.72,5.15) -- (7.12,0) -- cycle    ;

%Straight Lines [id:da0727882271401632] 
\draw    (240,201) -- (289.14,97.81) ;
\draw [shift={(290,96)}, rotate = 475.46] [fill={rgb, 255:red, 0; green, 0; blue, 0 }  ][line width=0.75]  [draw opacity=0] (10.72,-5.15) -- (0,0) -- (10.72,5.15) -- (7.12,0) -- cycle    ;

%Straight Lines [id:da02068180741540726] 
\draw    (240,96) -- (289.14,199.19) ;
\draw [shift={(290,201)}, rotate = 244.54000000000002] [fill={rgb, 255:red, 0; green, 0; blue, 0 }  ][line width=0.75]  [draw opacity=0] (10.72,-5.15) -- (0,0) -- (10.72,5.15) -- (7.12,0) -- cycle    ;

%Straight Lines [id:da4941096166166473] 
\draw    (340,96) -- (369.95,144.3) ;
\draw [shift={(371,146)}, rotate = 238.2] [fill={rgb, 255:red, 0; green, 0; blue, 0 }  ][line width=0.75]  [draw opacity=0] (10.72,-5.15) -- (0,0) -- (10.72,5.15) -- (7.12,0) -- cycle    ;

%Straight Lines [id:da0005753242144755921] 
\draw    (340,201) -- (370.02,147.74) ;
\draw [shift={(371,146)}, rotate = 479.41] [fill={rgb, 255:red, 0; green, 0; blue, 0 }  ][line width=0.75]  [draw opacity=0] (10.72,-5.15) -- (0,0) -- (10.72,5.15) -- (7.12,0) -- cycle    ;

%Straight Lines [id:da6663993224468625] 
\draw    (240,201) -- (288,201) ;
\draw [shift={(290,201)}, rotate = 180] [fill={rgb, 255:red, 0; green, 0; blue, 0 }  ][line width=0.75]  [draw opacity=0] (10.72,-5.15) -- (0,0) -- (10.72,5.15) -- (7.12,0) -- cycle    ;

%Straight Lines [id:da971612493075633] 
\draw    (240,96) -- (288,96) ;
\draw [shift={(290,96)}, rotate = 180] [fill={rgb, 255:red, 0; green, 0; blue, 0 }  ][line width=0.75]  [draw opacity=0] (10.72,-5.15) -- (0,0) -- (10.72,5.15) -- (7.12,0) -- cycle    ;

% Text Node
\draw (115,201) node [scale=1]  {$n_{0,2}$};
% Text Node
\draw (115,96) node [scale=1]  {$n_{0,1}$};
% Text Node
\draw (215,201) node [scale=1]  {$n_{1,2}$};
% Text Node
\draw (215,96) node [scale=1]  {$n_{1,1}$};
% Text Node
\draw (315,96) node [scale=1]  {$n_{2,1}$};
% Text Node
\draw (315,201) node [scale=1]  {$n_{2,2}$};
% Text Node
\draw (396,146) node [scale=1]  {$n_{3,1}$};
% Text Node
\draw (162.5,80) node [scale=1] [align=left] {1.9};
% Text Node
\draw (132.5,131) node [scale=1] [align=left] {1.1};
% Text Node
\draw (130,161) node [scale=1] [align=left] {\mbox{-}2.1};
% Text Node
\draw (161.5,212) node [scale=1] [align=left] {1.0};
% Text Node
\draw (264.5,80) node [scale=1] [align=left] {2.1};
% Text Node
\draw (238,131) node [scale=1] [align=left] {\mbox{-}0.9};
% Text Node
\draw (252.5,212) node [scale=1] [align=left] {1.1};
% Text Node
\draw (238,161) node [scale=1] [align=left] {\mbox{-}1.0};
% Text Node
\draw (372.5,110) node [scale=1] [align=left] {1.0};
% Text Node
\draw (375,182) node [scale=1] [align=left] {\mbox{-}1.0};
% Text Node
\draw (215,58)  node   {$\ReluOUT{n_{1,1}} =[ -2.2,9.3]$};
% Text Node
\draw (345,58)  node   {$\ReluOUT{n_{2,1}} =[ -4.7,11.8]$};
% Text Node
\draw (215,238) node   {$\ReluOUT{n_{1,2}} =[ 5.4,11.6]$};
% Text Node
\draw (345,238) node   {$\ReluOUT{n_{2,2}} =[ -.2,2.6]$};
% Text Node
\draw (52,83)   node   {$x_1 \in [ 4,6]$};
% Text Node
\draw (52,201)  node   {$x_2 \in [ 1,5]$};
% Text Node
\draw (50.5,103) node   {\textcolor{dkgreen}{$\ReluOUTdelta{x_1} =[ 0,0]$}};
% Text Node
\draw (50.5,220) node   {\textcolor{dkgreen}{$\ReluOUTdelta{x_2} =[ 0,0]$}};
% Text Node
\draw (215,40)   node   {\textcolor{dkgreen}{$\ReluOUTdelta{n_{1,1}} =[ 0,1.1]$}};
% Text Node
\draw (215,256)  node   {\textcolor{dkgreen}{$\ReluOUTdelta{n_{1,2}} =[ -.6,-.4]$}};
% Text Node
\draw (345,42)   node   {\textcolor{dkgreen}{$\ReluOUTdelta{n_{2,1}} =[ -.53,3.02]$}};
% Text Node
\draw (345,256)  node   {\textcolor{dkgreen}{$\ReluOUTdelta{n_{2,2}} =[ -3.8,0]$}};
% Text Node
\draw (430,102) node   {\textcolor{dkgreen}{$\ReluOUTdelta{n_{3,1}} =[ -.53,6.81]$}};

\end{tikzpicture}}
\caption{Forward interval analysis of a neural network.} 
\label{fig:forward-pass-example}
\end{figure}
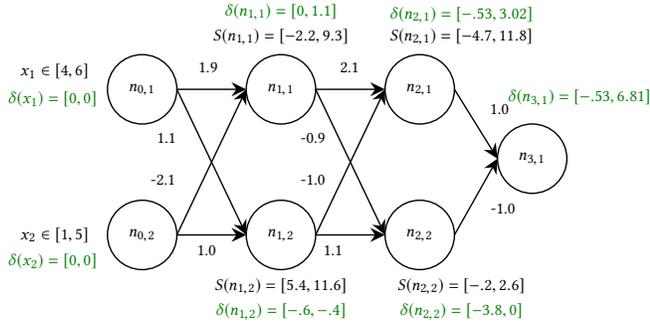

As a result, our method can solve the differential verification
problems much more efficiently.  Consider the litmus test of verifying the
equivalence of two identical networks.  Our method can obtain a formal
proof (that $|f'-f|<\epsilon$) after performing the forward interval
analysis once; in contrast, all other existing tools have failed to do
so.
For the example in Figure~\ref{fig:forward-pass-example}, we can prove
the output difference $\ReluOUTdelta{n_{3,1}}$ is bounded by $[-0.53,
6.81]$ after only the first pass.
It also outperforms existing tools on other verification problems
where $f'$ is obtained from $f$ through quantization; details of the
experimental comparisons are in Section~\ref{sec:experiment}.

\section{Preliminaries}
\label{sec:prelims}

First, we review the basics of interval analysis for neural networks.

\subsection{Neural Networks}

We consider a neural network as a non-linear function that takes some
value in $\mathbb{R}^n$ as input and returns some value in
$\mathbb{R}^m$ as output, where $n$ is the number of input variables
and $m$ is the number of output variables.
Let the network $ f $ be denoted $f: \mathbb{X} \to \mathbb{Y}$, where
$\mathbb{X} \subseteq \mathbb{R}^n$ is the input domain and
$ \mathbb{Y} \subseteq \mathbb{R}^m $ is the output domain.
In image recognition applications, for instance, $\mathbb{X}$ may be a vector of pixels representing an image and
$\mathbb{Y}$ may be a vector of probabilities for class labels.  In
aircraft collision detection, on the other hand, $\mathbb{X}$ may be
sensor data and $\mathbb{Y}$ may be a set of actions to take.

In this work, we consider fully-connected feed-forward networks with
rectified linear unit (ReLU) activations, which are the most popular
in practical hardware/software implementations.
Thus, $y = f(x)$ is a series of affine transformations (e.g., $x \cdot
W_1$ = $\Sigma_i x_i w_{1,i}$) followed by point-wise ReLU (e.g.,
$\ReLU{x\cdot W_1} = max(0, x\cdot W_1)$).
Let $W_k$, where $1 \leq k \leq l$, be the weight matrix associated
with the $k$-th layer, and $l$ be the number of layers; the affine
transformation in the $k$-th layer is a standard matrix
multiplication, followed by the point-wise application of ReLU.

Formally, $ f = f_l( f_{l-1} (... f_2( f_1(x \cdot W_1) \cdot W_2 ))
... \cdot W_{l-1}) $, where each $ f_k $, $1\leq k \leq l$, is a
point-wise ReLU.
For the network in Figure~\ref{fig:single-network}, in particular, the
input is a vector $x = \{x_1,x_2\}$, the weight matrix $W_3 = \{ 1.0,
-1.0 \}^T$, and $x \cdot W_1 = \{ 1.9 x_1 -2.1 x_2, 1.1 x_1 + 1.0
x_2 \}$.

For ease of presentation, we denote the weight of the edge from
the $i$-th neuron of layer $ k - 1 $ to the $j$-th neuron of layer $ k
$ as $ W_k[i,j] $. We also denote the $j$-th neuron of layer $ k $ as
$ n_{k,j} $.

\subsection{Interval Analysis}
\label{sec:int_analysis}
To ensure that our analysis is over-approximated, we use interval
analysis~\cite{moore2009introduction}, which can be viewed as a
specific instantiation of the general abstract
interpretation~\cite{CousotC77} framework.  
%It is capable of
%approximating the output of a mathematical function over a defined
%input interval region. 
Interval analysis is well-suited for analyzing ReLU neural networks as
it has well-defined transformers over addition, subtraction, and
scaling (i.e., multiplication by a constant).

Interval addition as denoted $[a,b] + [c,d] = [a+c, b+d]$
does not lead to loss of accuracy. 
Scaling as denoted $[a,b] * c = [a*c, b*c]$ when $c\geq 0$, or
$[a,b]*c = [b*c, a*c]$ when $c<0$, does not lead to loss of accuracy either.
Interval subtraction as denoted $[a,b] - [c,d] = [a-d, b-c]$, however,
may lead to accuracy loss.

To illustrate, consider $ f(x) = 2.1x $ and $ f'(x) = 2x $, and say we
want to approximate their difference for the input region $ x\in [-1,1] $. Using
interval arithmetic, we would compute $ f([-1,1]) - f'([-1,1]) =
[-2.1,2.1] - [-2,2] = [-4.1, 4.1] $. Clearly this is far from the
exact interval of $f(x) -f'(x) = 2.1x - 2x = 0.1x$ over $x\in[-1,1]$,
which is $ [-0.1, 0.1] $.
The reason for such loss of accuracy is that, during interval
arithmetic, the relationship between values of $2.1x$ and $2x$ (i.e.,
they are for the same value of $x$) is lost.

\subsection{Symbolic Interval}

One way to overcome the accuracy loss is using \textit{symbolic
intervals}~\cite{WangPWYJ18}, which can encode the constraint that
inputs to $ f $ and $ f' $ are actually related.  With this technique,
we would use the symbol $ x $ with the constraint $ x \in [-1,1] $ to
initialize the input intervals. Then, the computation becomes $ f([x,
x]) - f'([x,x]) = [2.1x, 2.1x] - [2x, 2x] = [0.1x, 0.1x] $. Finally,
we would compute the upper and lower bounds for $x\in [-1,1]$ and
return the precise interval $ [-0.1, 0.1] $.

Unfortunately, symbolic intervals depend on $ f $ and $ f' $ being
linear in the entire input region in order to be sound.  Indeed, if we
add ReLU to the functions, i.e., $\ReLU{ f(x) } = max(0,2.1x) $
and $\ReLU{ f'(x) } = max(0,2x) $, where $x\in [-1,1]$, then the
lower and upper bounds are no longer precise nor sound.
\textcolor{black}{
The reason is because $max(0, 2.1x)$ is non-linear in $x\in [-1,1]$.
Thus, we have to approximate using the concrete interval $[0,2.1]$.
Similarly, $max(0,2x)$ is approximated using $[0,2]$.  Thus, $[0, 2.1]
- [0, 2] = [-2, 2.1]$.
}

\subsection{Refinement}

To improve the accuracy of the symbolic interval analysis, we need to
divide the input region into subregions.  The intuition is that,
within a smaller subregion, the ReLU is less likely to exhibit
non-linear behavior and force the analyzer to over-approximate.
Consider $\ReLU{2.1x} = max(0, 2.1x)$, where $x\in
[-1,1]$. After the input region is divided into subregions $x\in
[-1,0] \cup [0,1]$, we have $max(0,2.1x) = [0,0]$ for $x\in[-1,0]$ and
$max(0,2.1x) = [2.1x,2.1x]$ for $x\in[0,1]$.  In both cases, the
intervals are precise -- there is no approximation at all.

When we only have one input variable, we do not have a choice on which
variable to refine. However, neural networks have many inputs, and
refining some of them will not always yield benefit. Thus, we have to
identify the right input to split.

Consider $f(x_1,x_2) = \ReLU{5x_1} + \ReLU{2x_2}
- \ReLU{x_2}$, where $x_1\in [1,3]$ and $x_2\in [-1,1]$.
The initial analysis is not very accurate due to approximations caused
by the ReLU: 
$\ReLU{5 * [1,3]} + \ReLU{2* [-1,1]} + \ReLU{ [-1,1] }$
$=  \ReLU{[5,15]} + \ReLU{[-2,2]} + \ReLU{[-1,1]}$
$=  [5,15] + [0,2] - [0,1]$
$=  [4,17]$.

\begin{comment}
To compute the refinement, \ReluVal{} would compute the partial
derivative $\partial f/\partial x_1$, represented as an interval gradient $g_{x_1} =
[5,5]$.  Similarly, $\partial f/\partial x_2$ is an interval gradient $g_{x_2} = [0,2]
- [0,1] = [-1,2]$.  The width of the input ranges are $wd_{x_1} = 3-1
= 2$ and $wd_{x_2} = 1 - (-1) = 2$.  Therefore, the smear values are
computed as follows:
%
$ \UB{g_{x_1}} \times wd_{x_1} =  5 \times 2 = 10$ and
$ \UB{g_{x_2}} \times wd_{x_2} =  2 \times 2 = 8$.
\end{comment}

If we split $x_1 \in [1,3]$ into $x_1\in [1,2] \cup [2,3]$ and
perform interval analysis for both subregions, the output would be
$[4,12]\cup[9,17] = [4,17]$, which does not improve over the initial
result.
%
\begin{comment}
\[\begin{array}{rl}
FwdPass\#2: & \mbox{ splitting variable } x_1 \\
       x\in [1,2]&    \proc{ReLu}(5 * [1,2]]) + \proc{ReLu}(2* [-1,1]) + \proc{ReLu}( [-1,1] ) \\
                 & =  \proc{ReLu}([5,10]) + \proc{ReLu}([-2,2]) + \proc{ReLu}([-1,1]) \\
                 & =  [5,10] + [0,2] - [0,1]\\
                 & =  [5,10] + [-1,2]\\
                 & =  [4,12]\\
       x\in [2,3]&    \proc{ReLu}(5 * [2,3]]) + \proc{ReLu}(2* [-1,1]) + \proc{ReLu}( [-1,1] ) \\
                 & =  \proc{ReLu}([10,15]) + \proc{ReLu}([-2,2]) + \proc{ReLu}([-1,1]) \\
                 & =  [10,15] + [0,2] - [0,1]\\
                 & =  [10,15] + [-1,2]\\
                 & =  [9,17]\\
        combined &  [4,12] \cup [9,17] = [4,17]   \mbox{ no improvement }\\
\end{array}\]
\end{comment}

In contrast, if we split $x_2\in[-1,1]$ into $x_2\in[-1,0] \cup
[0,1]$, the accuracy would improve significantly.
%
\begin{comment}
\[\begin{array}{rl}
FwdPass\#2: & \mbox{ splitting variable } x_2 \\
      y\in [-1,0]&    \proc{ReLu}(5 * [1,3]]) + \proc{ReLu}(2* [-1,0]) + \proc{ReLu}( [-1,0] ) \\
                 & =  \proc{ReLu}([5,15]) + \proc{ReLu}([-2,0]) + \proc{ReLu}([-1,0]) \\
                 & =  [5,15] + [0,0] - [0,0]\\
                 & =  [5,15]\\
       y\in [0,1]&    \proc{ReLu}(5 * [1,3]]) + \proc{ReLu}(2* [0,1]) + \proc{ReLu}( [0,1] ) \\
                 & =  \proc{ReLu}([5,15]) + \proc{ReLu}([0,2]) + \proc{ReLu}([0,1]) \\
                 & =  [5,15] + [2y,2y] - [y,y]\\
                 & =  [5,15] + [y,y]\\
                 & =  [5,16]\\
        combined &  [5,15] \cup [5,16] = [5,16]   \mbox{ improved }\\
\end{array}
\]
\end{comment}
%
Since the ReLU is always activated for $x_2\in[0,1]$,
$\ReLU{2x_2}$ and $\ReLU{x_2}$ can be represented by
$[2x_2,2x_2]$ and $[x_2,x_2]$, respectively, and $\ReLU{2x_2}
- \ReLU{x_2} = [x_2,x_2] = [0,1]$.
Since the ReLU is always de-activated for $x_2\in[-1,0]$, we have
$\ReLU{2x_2} - \ReLU{x_2} = [0,0]$.
Thus, the combined output $[5,15]\cup[5,16] = [5,16]$ is more accurate
than the initial approximation $[4,17]$.

While how to analyze non-linear activation functions such as ReLU has
been studied in prior work~\cite{WangPWYJ18nips, SinghGPV19, KatzHIJLLSTWZDK19}, none of the existing techniques touch
upon the complex scenarios arising from differential verification of
two closely related networks. Our work fills the gap.  Specifically,
we propose a more accurate forward pass for the interval analysis
(Section~\ref{sec:forward-pass}) and a more accurate backward pass for
the refinement (Section~\ref{sec:backward-pass}).

\begin{comment}%chao: too complex for now...
\begin{figure*}
	\scalebox{1}{
		\input{figs/relu_cases.tex}	
	}
	\caption{The nine cases for the ReLU transformation.\label{fig:relu_cases}}
\end{figure*}
\end{comment}

\begin{figure}
	\scalebox{0.8}{
	\tikzset{every picture/.style={line width=0.75pt}} %set default line width to 0.75pt        

\begin{tikzpicture}[x=0.75pt,y=0.75pt,yscale=-1,xscale=1]
%uncomment if require: \path (0,300); %set diagram left start at 0, and has height of 300

%Shape: Circle [id:dp13857136549466964] 
\draw   (100,90) .. controls (100,78.95) and (108.95,70) .. (120,70) .. controls (131.05,70) and (140,78.95) .. (140,90) .. controls (140,101.05) and (131.05,110) .. (120,110) .. controls (108.95,110) and (100,101.05) .. (100,90) -- cycle ;
%Shape: Circle [id:dp8027912343340515] 
\draw   (220,130) .. controls (220,118.95) and (228.95,110) .. (240,110) .. controls (251.05,110) and (260,118.95) .. (260,130) .. controls (260,141.05) and (251.05,150) .. (240,150) .. controls (228.95,150) and (220,141.05) .. (220,130) -- cycle ;
%Shape: Circle [id:dp6531860577539721] 
\draw   (100,160) .. controls (100,148.95) and (108.95,140) .. (120,140) .. controls (131.05,140) and (140,148.95) .. (140,160) .. controls (140,171.05) and (131.05,180) .. (120,180) .. controls (108.95,180) and (100,171.05) .. (100,160) -- cycle ;
%Straight Lines [id:da3593349023905198] 
\draw    (140,90) -- (220.18,118.98) ;
\draw [shift={(223,120)}, rotate = 199.87] [fill={rgb, 255:red, 0; green, 0; blue, 0 }  ][line width=0.08]  [draw opacity=0] (10.72,-5.15) -- (0,0) -- (10.72,5.15) -- (7.12,0) -- cycle    ;

%Straight Lines [id:da8904753170585369] 
\draw    (140,160) -- (220.08,140.7) ;
\draw [shift={(223,140)}, rotate = 526.45] [fill={rgb, 255:red, 0; green, 0; blue, 0 }  ][line width=0.08]  [draw opacity=0] (10.72,-5.15) -- (0,0) -- (10.72,5.15) -- (7.12,0) -- cycle    ;

%Straight Lines [id:da007203855246534552] 
\draw    (140,120) -- (217.01,126.74) ;
\draw [shift={(220,127)}, rotate = 185] [fill={rgb, 255:red, 0; green, 0; blue, 0 }  ][line width=0.08]  [draw opacity=0] (10.72,-5.15) -- (0,0) -- (10.72,5.15) -- (7.12,0) -- cycle    ;

%Straight Lines [id:da72260506902163] 
\draw    (140,134) -- (217,130.15) ;
\draw [shift={(220,130)}, rotate = 537.14] [fill={rgb, 255:red, 0; green, 0; blue, 0 }  ][line width=0.08]  [draw opacity=0] (10.72,-5.15) -- (0,0) -- (10.72,5.15) -- (7.12,0) -- cycle    ;

% Text Node
\draw (120,125) node  [rotate=-90] [align=left] {. . .};
% Text Node
\draw (120,90) node    {$n_{k-1,1}$};
% Text Node
\draw (240,130) node    {$n_{k,j}$};
% Text Node
\draw (120,160) node    {$n_{k-1,i}$};
% Text Node
\draw (199.5,91) node  [font=\Large]  {$W_{k}[ 1,j]$};
% Text Node
\draw (199.5,164) node  [font=\Large]  {$W_{k}[ i,j]$};
% Text Node
\draw (127.5,60) node    {$layer\ k-1$};
% Text Node
\draw (245.5,60) node    {$layer\ k$};

\end{tikzpicture}
	}
	\caption{Diagram of weight notations.\label{fig:neuron_weights}}
\end{figure}
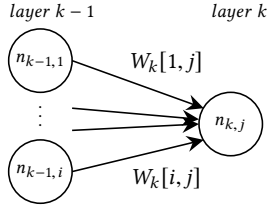

\section{Forward Interval Analysis}
\label{sec:forward-pass}

In this section, we describe our forward pass for computing the value
differences between neurons in the two networks. 
Recall that network $f$ has $l$ layers and weight matrices $W_k$,
$1 \leq k \leq l$, and $n_{k,j}$ is the $j$-th node in the $k$-th
layer. Furthermore, $W_k[i,j]$ is the weight of the edge from
$n_{{k-1},i}$ to $n_{k,j}$. We illustrate these notations in Figure~\ref{fig:neuron_weights}.
Similarly, network $f'$ has weight matrices $ W'_k$, nodes $n'_{k,j}$,
and weights $W'_k[i,j]$ .
Let $ W^\Delta_k[i,j] $ be the weight difference, i.e $ W^\Delta_k[i,j] = W'_k[i,j]-W_k[i,j] $.

We now define notations for the interval values of neurons.
Since each neuron $n_{k,j}$ has an affine transformation (multiplying by the incoming weights)
and a ReLU, we denote the input interval to the neuron (after applying the affine transform) as $\ReluIN{n_{k,j}}$,
and we denote the output interval of the neuron (after applying
the ReLU) as $\ReluOUT{n_{k,j}}$.
\ignore{Similarly, $\ReluIN{n'_{k,j}}$ is the input of the ReLU 
and $\ReluOUT{n'_{k,j}}$ is the ReLU output.}
We denote the interval bound on the difference between the inputs to 
$ \np $ and $ \n $ as
$\ReluINdelta{n_{k,j}}$, and we denote the interval
difference between the outputs as $\ReluOUTdelta{n_{k,j}}$.
Finally, we denote the symbolic upper and lower bound of any value using the notation
$\UB{}$ and $\LB{}$. For example, $ \UB{\ReluOUT{n_{k,j}}} $ and $ \LB{\ReluOUT{n_{k,j}}}$
denote the symbolic upper and lower bound for the output of neuron $ n_{k,j} $.

With this notation, our forward pass is shown in
Algorithms~\ref{alg:fwdpass} and~\ref{alg:relu}.  The input consists of the two networks,
$f$ and $f'$, and the input region of interest $ X $, which defines 
an interval for each input neuron.
After initializing the input intervals, the algorithm iteratively computes each
$\ReluOUT{n_{k,j}}, \ReluOUT{n'_{k,j}}$, and $ \ReluOUTdelta{n_{k,j}}$ of the
subsequent layer by applying the affine transformation followed by the ReLU
transformation. The algorithm iterates until the output layer is reached.
In addition, it computes the \textit{gradient masks} for the neurons of $ f $ and $ f' $, denoted as $ R $ and $ R' $,
which record the state of each neuron in the forward pass ($ [0,0] $ is inactive, $ [1,1] $ is active, and $ [0,1] $ is both).
These are used in the refinement phase (Section~\ref{sec:backward-pass}) to
determine which input neuron to refine.

We omit discussion of computing $ \ReluIN{n_{k,j}} $ and $ \ReluOUT{n_{k,j}} $
because it has been studied in previous work~\cite{WangPWYJ18}. We focus on computing
$ \ReluINdelta{n_{k,j}} $ and $ \ReluOUTdelta{n_{k,j}} $ in the following section.

\begin{algorithm}[t]
\caption{Forward symbolic interval analysis.}
\label{alg:fwdpass}
{\footnotesize
	\SetAlgoLined
	\KwIn{ network $f$, network $f'$, input region $X$}
	\KwResult{ difference $\{ \ReluOUTdelta{n_{l,j}} \}$ for output }

	Initialize $\{ \ReluOUT{n_{0,j}} \}, \{ \ReluOUT{n'_{0,j}} \}$ to input region $X$ and 
                   $\{ \ReluOUTdelta{n_{0,j}} \}$ to  $0$\\

	\For{k \textbf{in} 1..$N_{Layer}$}{

		\textcolor{dkgreen}{// Affine transformer} \\
                \For{j \textbf{in} 1..layerSize[k]}{
                   $\ReluIN{ n_{k,j}} \leftarrow \sum_i  \ReluOUT{n_{k-1,i} } \cdot  W_k[i,j]$\\
                   $\ReluIN{n'_{k,j}} \leftarrow \sum_i  \ReluOUT{n'_{k-1,i}} \cdot W'_k[i,j]$\\
  		   		   $ \ReluINdelta{n_{k,j}} \leftarrow  \sum_i ( \ReluOUT{n_{k-1,i}} \cdot W^\Delta_k[i,j] 
                                                        +  \ReluOUTdelta{n_{k-1,i}} \cdot W'_k[i,j] )$\\
                   %$\ReluIN{n'_{k,j}} \leftarrow \ReluIN{n_{k,j}} + \ReluINdelta{n_{k,j}}$\\
                }

		\If{k = $N_{Layer}$}{
		    \Return \{$\ReluINdelta{n_{k,j}}$\}  
		}

		\textcolor{dkgreen}{// ReLU transformer} \\
                \For{j \textbf{in} 1..layerSize[k]}{
                   $\langle \ReluOUT{n_{k,j}}, \ReluOUT{n'_{k,j}},  \ReluOUTdelta{n_{k,j}} \rangle   \leftarrow$
                   \proc{ReLuTransform}$( \ReluIN{n_{k,j}}, \ReluIN{n'_{k,j}},  \ReluINdelta{n_{k,j}} )$\\
                }
	}
}
\end{algorithm}

\subsection{The Affine Transformer}

Computing $\ReluINdelta{n_{k,j}}$ involves two steps.
First, we compute $\EdgeOUTdelta{W_k[i,j]}$ for each incoming edge to
$n_{k,j}$. Here, $\EdgeOUTdelta{W_k[i,j]}$ is the difference in values produced
by the edges from $n_{k-1,i}$ to $n_{k,j}$ and from $n'_{k-1,i}$ to
$n'_{k,j}$.
Second, we sum them to obtain $\ReluINdelta{n_{k,j}}$.

In the first step, there are two components to consider when computing
$\EdgeOUTdelta{W_k[i,j]}$.
First, there is the ``new quantity''
introduced by the difference in edge weights, which formally is 
$\ReluOUT{n_{k-1,i}} \cdot W^\Delta_k[i,j]$. In English,
this is the interval of neuron $n_{k-1,i}$ in the previous layer multiplied by the edge
weight difference in the current layer.
Second, there is the ``old quantity'' accumulated in previous
layers being scaled by the edge weight in the current layer. Formally this is
$\ReluOUTdelta{n_{k-1,i}} \cdot W'_k[i,j]$. Below we write out the formal 
derivation:
\begin{align*}
 \EdgeOUTdelta{W_k[i,j]} &= W'_k[i,j] \cdot S(n'_{k-1,i}) - W_k[i,j] \cdot S(n_{k-1, i}) \\
 			&= W'_k[i,j] \cdot S(n'_{k-1,i}) - W_k[i,j] \cdot S(n_{k-1, i}) + \\
			&   (\textcolor{red}{W'_k[i,j] \cdot S(n_{k-1, i})} \textcolor{red}{- W'_k[i,j] \cdot S(n_{k-1, i})}) \\
 			&=  (W'_k[i,j] \cdot S(n'_{k-1,i}) \textcolor{red}{- W'_k[i,j] \cdot S(n_{k-1, i})}) + \\
			&   (\textcolor{red}{W'_k[i,j] \cdot S(n_{k-1, i})} - W_k[i,j] \cdot S(n_{k-1, i})) \\
 			&= \ReluOUTdelta{n_{k-1,i}} \cdot W'_k[i,j] + \ReluOUT{n_{k-1,i}} \cdot W^\Delta_k[i,j]
                            ~.
\end{align*}

In the second step, we sum together each incoming 
$\EdgeOUTdelta{W_k[i,j]}$ term to obtain $\ReluINdelta{n_{k,j}}$,
which is the difference of the values $\ReluIN{n_{k,j}}$ and
$\ReluIN{n'_{k,j}}$.  That is,
\[
  \ReluINdelta{n_{k,j}} = \sum_i \EdgeOUTdelta{W_k[i,j]}   ~.
\]

We demonstrate the computation on the example in
Figure~\ref{fig:forward-pass-example}.  First, we compute
$\EdgeOUTdelta{W_1[1,1]} = 0.1 \cdot [4,6] + 2 \cdot [0,0] = [0.4,
0.6] $ and $\EdgeOUTdelta{W_1[2,1]} = 0.1 \cdot [1,5] + 2 \cdot [0,0]
= [0.1,0.5] $.
Then, we compute 
$\EdgeOUTdelta{W_1[1,2]} = [-0.6, -0.4] $ and 
$\EdgeOUTdelta{W_1[2,2]} = [0,0] $.

Next, we compute 
$\ReluINdelta{n_{1,1}} = \EdgeOUTdelta{W_1[1,1]} + \EdgeOUTdelta{W_1[2,1]} =$ \\$ [0.5, 1.1] $ and 
$\ReluINdelta{n_{1,2}} = \EdgeOUTdelta{W_1[1,2]} + \EdgeOUTdelta{W_1[2,2]} = [-0.6, -0.4] $.

\subsection{The ReLU Transformer}

Next, we apply the ReLU activation to $\ReluINdelta{n_{k,j}}$ to obtain
$\ReluOUTdelta{n_{k,j}}$.
We consider  nine cases
based on whether the ReLUs of $n_{k,j}$ and $n'_{k,j}$
are \emph{always activated}, \emph{always deactivated},
or \emph{non-linear}.
In the remainder of the section, we discuss how to soundly
over-approximate. Algorithm~\ref{alg:relu} shows the details.

\begin{algorithm}[t]
\caption{Over-approximating ReLU activation function.}
\label{alg:relu}
{\footnotesize

	\SetAlgoLined
	\KwIn{ value $ \ReluIN{n_{k,j}}$, value $\ReluIN{n'_{k,j}}$, difference $\ReluINdelta{n_{k,j}}$ }
	\KwResult{ value $\ReluOUT{n_{k,j}}$, value $\ReluOUT{n'_{k,j}}$, difference $\ReluOUTdelta{n_{k,j}}$ }
        
        \uIf{$ \UB{\ReluIN{n_{k,j}}} \leq 0$}{

				 $R [k][j] = [0,0]$\\
                 $\ReluOUT{n_{k,j}} = [ 0, 0 ] $

                 \uIf{ $\UB{\ReluIN{n'_{k,j}}} \leq 0$}{
                                $R'[k][j] = [0,0]$\\
                                $\ReluOUT{n'_{k,j}} = [ 0, 0 ] $\\

                                $\ReluOUTdelta{n_{k,j}}  = [0, 0]$
                 } \uElseIf{$\LB{\ReluIN{n'_{k,j}}} >0$} {
                                $R'[k][j] = [1,1]$\\
                                $\ReluOUT{n'_{k,j}} = \ReluIN{n'_{k,j}}$\\

                                $\ReluOUTdelta{n_{k,j}}  =  \ReluIN{n'_{k,j}}$
                 } \uElse {
                                $R'[k][j] = [0,1]$\\
                                $\ReluOUT{n'_{k,j}} = [ 0, \UBconcrete{\ReluIN{n'_{k,j}}} ]$\\

                                $\ReluOUTdelta{n'_{k,j}} = [ 0, \UBconcrete{\ReluIN{n'_{k,j}}} ]$
                 }
			
        } \uElseIf{$ \LB{\ReluIN{n_{k,j}}} > 0 $}{

                 $R [k][j] = [1,1]$\\
                 $\ReluOUT{n_{k,j}} = \ReluIN{n_{k,j}}$ 

                 \uIf{ $\UB{\ReluIN{n'_{k,j}}} \leq 0$}{
                                $R'[k][j] = [0,0]$\\
                                $\ReluOUT{n'_{k,j}} = [ 0, 0 ] $\\

                                $\ReluOUTdelta{n_{k,j}} = - \ReluIN{n_{k,j}}$\\ 
                 } \uElseIf{$\LB{\ReluIN{n'_{k,j}}} >0$} {
                                $R'[k][j] = [1,1]$\\
                                $\ReluOUT{n'_{k,j}} = \ReluIN{n'_{k,j}}$ 

                                $\ReluOUTdelta{n_{k,j}} = \ReluINdelta{n_{k,j}}$\\
                 } \uElse {
                                $R'[k][j] = [0,1]$\\
                                $\ReluOUT{n'_{k,j}} = [ 0, \UBconcrete{\ReluIN{n'_{k,j}}} ]$\\

                                \textcolor{black}{$\ReluOUTdelta{n_{k,j}} = [-\UB{\ReluIN{n_{k,j}}}, 
                                                          \UBconcrete{\ReluIN{n'_{k,j}}}- \LB{\ReluIN{n_{k,j}}}]$}\\

\textcolor{dkgreen}{		Opt. 1: \\
                                $\ReluOUTdelta{n_{k,j}} = [ 
                                      max\big( - \UBconcrete{\ReluIN{n_{k,j}}}, \LBconcrete{\ReluINdelta{n_{k,j}}} \big)$,\\
                                      $\qquad \qquad \;\; max\big( -\LBconcrete{\ReluIN{n_{k,j}}}, \UBconcrete{\ReluINdelta{n_{k,j}}} \big) ]$\\
}
                 }
			
	} \uElse{

                 $R [k][j] = [0,1]$\\
                 $\ReluOUT{n_{k,j}} = [ 0, \UBconcrete{\ReluIN{n_{k,j}}} ]$

                 \uIf{ $\UB{\ReluIN{n'_{k,j}}} \leq 0$}{
                                $R'[k][j] = [0,0]$\\
                                $\ReluOUT{n'_{k,j}} = [ 0, 0 ] $\\

                                $\ReluOUTdelta{n_{k,j}} = [ - \UBconcrete{\ReluIN{n_{k,j}}},  0 ]$
                 } \uElseIf{$\LB{\ReluIN{n'_{k,j}}} >0$} {
                                $R'[k][j] = [1,1]$\\
                                $\ReluOUT{n'_{k,j}} = \ReluIN{n'_{k,j}}$ 

                                \textcolor{black}{$\ReluOUTdelta{n_{k,j}} = [ \LB{\ReluIN{n'_{k,j}}}-\UBconcrete{\ReluIN{n_{k,j}}}, \UB{\ReluIN{n'_{k,j}}} ]$}\\
\textcolor{dkgreen}{            Opt. 2: \\
				$\ReluOUTdelta{n_{k,j}} = [ min\big(\LBconcrete{\ReluIN{n'_{k,j}}}, \LBconcrete{\ReluINdelta{n_{k,j}}} \big),$\\  
                                           $\qquad \qquad \;\;\; min\big(\UBconcrete{\ReluINdelta{n_{k,j}}}, \UBconcrete{\ReluIN{n'_{k,j}}} \big) ] $\\
                                }
                                                           
                 } \uElse {
                                $R'[k][j] = [0,1]$\\
                                $\ReluOUT{n'_{k,j}} = [ 0, \UBconcrete{\ReluIN{n'_{k,j}}} ]$\\
\textcolor{black}{
                                $\ReluOUTdelta{n_{k,j}} =  [-\UBconcrete{\ReluIN{n_{k,j}}},
                                                            \UBconcrete{\ReluIN{n'_{k,j}}}]$}\\

\textcolor{dkgreen}{		Opt. 3: \\ 
                                \uIf{$\LB{\ReluINdelta{n_{k,j}}} \geq 0$}{
                                     $\ReluOUTdelta{n_{k,j}} = [ 0, \UBconcrete{\ReluINdelta{n_{k,j}}} ]$
                                } \uElseIf{$ \UB{\ReluINdelta{n_{k,j}}} \leq 0 $} {
                            		$\ReluOUTdelta{n_{k,j}} = [max\big(\LBconcrete{\ReluINdelta{n_{k,j}}}, -\UBconcrete{\ReluIN{\n}}\big), 0 ]$
                        		} \uElse {
                                    $\ReluOUTdelta{n_{k,j}} = [ max\big(  \LBconcrete{\ReluINdelta{n_{k,j}}}, 
 								     -\UBconcrete{\ReluIN{n_{k,j}}} \big)$, \\
                                          $ \qquad \qquad \;\;\; \UBconcrete{\ReluIN{n'_{k,j}}} ]$
                                }
                            }
                 }
         }
}

\end{algorithm}

First, we consider the three cases when the ReLU in $\n$ is
\emph{always deactivated}.  
(1) If the ReLU in $ \np $ is also \emph{always deactivated}, the
outputs of both ReLUs are 0, and the difference is 0.
(2) If the ReLU in $ \np $ is \emph{always activated}, the output
difference will be $\ReluIN{\np} - [0,0] = \ReluIN{\np}$.  Note that
we can maintain symbolic equations here.
(3) If the ReLU in $ \np $ is \emph{non-linear}, the 
difference will be $[0, \UBconcrete{\ReluIN{\np}}] $.
While $\UB{\ReluIN{\np}}$ is the symbolic upper bound of
$\ReluIN{\np}$, $\UBconcrete{\ReluIN{\np}}$ is the
concrete upper bound. Note that since $ \np $ is non-linear,
we must concretize to be sound.

Next, we consider the three cases when the ReLU in $\n$ is
\emph{always activated}. 
(1) If the ReLU in $\np$ is \emph{always deactivated},  the
difference is $[0,0] - \ReluIN{\n}$. Again, we can soundly
maintain symbolic equations.
(2) If the ReLU in $\np$ is \emph{always activated}, then the
difference is the same as $\ReluINdelta{\n}$.
(3) If the ReLU in $\np$ is \emph{non-linear},  the
difference is $[0, \UBconcrete{\ReluIN{\np}}] -
\ReluIN{\n}$, which is  the same as 
$[-\UB{\ReluIN{\n}}, \UBconcrete{\ReluIN{\np}}
- \LB{\ReluIN{\n}}]$. Again, we concretize to ensure
soundness.

Third, we consider the three cases where the ReLU in $\n$
is \emph{non-linear}. 
(1) If the ReLU in $ \np $ is \emph{always deactivated}, the
difference is $[0,0] - [0, \UBconcrete{\ReluIN{\n}}]$. 
(2) If the ReLU in $ \np $ is \emph{always activated}, the
difference is $\ReluIN{\np} - [0,
\UBconcrete{\ReluIN{\n}}]$, which is the same as 
$[ \LB{\ReluIN{\np}}-\UBconcrete{\ReluIN{\n}}, \UB{\ReluIN{\np}}
]$.
(3) If the ReLU in $\np$ is also \emph{non-linear}, then the
difference is $[0, \UBconcrete{\ReluIN{\np}}] - [0,
\UBconcrete{\ReluIN{\n}}]$, which is  $[-\UBconcrete{\ReluIN{\n}},
\UBconcrete{\ReluIN{\np}}]$.

\subsection{Optimization}
\label{sec:opt}
\newcommand{\R}[1]{\proc{ReLU}\big(#1\big)}
The most important optimization we make in the forward pass when computing 
$ \ReluOUTdelta{\n} $ is in shifting from bounding the equation
\[
\R{\ReluIN{\np}} - \R{\ReluIN{\n}}
\]
to bounding one the following \textit{equivalent} equations
\begin{align}
\R{\ReluIN{\n} + \ReluINdelta{\n}} - \R{\ReluIN{\n}} \label{eq:1}\\
\R{\ReluIN{\np}} - \R{\ReluIN{\np} - \ReluINdelta{\n}} \label{eq:2}
\end{align}
Equation~\ref{eq:1} says that for any concrete $ \n \in \ReluIN{\n} $
, the most $ \np $ can change is bounded by $ \ReluINdelta{\n} $,
and similarly for Equation~\ref{eq:2}.
As shown in Algorithm~\ref{alg:relu}, we have identified three
optimization opportunities, marked as \emph{Opt.1-3}.\ignore{\footnote{
We note that the bounds from Equations~\ref{eq:1} and~\ref{eq:2} can actually be soundly applied
in all optimization cases. However, at the
time of submission, we had not yet proved this to be sound. This is
why we only apply the bounds from Equation~\ref{eq:1} to opt. 1, etc...\label{fn:1}.
See our arXiv paper~\cite{arxivPaper} for the most accurate version of the forward pass.}} We note that, 
even though we widen the difference
interval in some of these cases, the interval is almost always tighter
than if we subtract the bounds of $ \np $ and $ \n $, even when they
are symbolic. Below, we give formal proofs for most of the bounds,
and the remaining proofs can be found in the appendix
of our arXiv paper~\cite{arxivPaper}.

\subsubsection{Opt. 1: $ \n $ is active, $ \np $ is non-linear.}
\label{sec:opt1}
Using Equation~\ref{eq:1}, we can potentially tighten the
lower bound. To reduce the notation complexity, we rewrite Equation~\ref{eq:1} as the function:
\begin{align}
\deq(n, d) & = \ReLU{n + d} - \ReLU{n} \label{eq:3}\\
        & = \ReLU{n + d} - n \label{eq:4}
\end{align}
where $n \in \ReluIN{\n}$ and $d \in \ReluINdelta{\n}$, and we can simplify
from Equation~\ref{eq:3} to~\ref{eq:4} because $ \n $ is active. Now, computing
$ \ReluOUTdelta{\n} $ amounts to finding the upper and lower bounds on $ \deq(n,d) $.

Observe that if $ n + d \geq 0 $ then $ \deq(n, d) = d $ because $ \proc{ReLU}(n + d)$ 
simplifies to $ n + d $, and the like terms cancel.
Otherwise $ \deq(n, d) = -n$ because $ \proc{ReLU}(n + d) = 0 $.
Observing that $ n + d \geq 0 = d \geq -n $, this means $ \deq(n,d) $
is equivalent to:
\[
\deq(n, d) = \begin{cases}
d & d \geq -n\\
-n & d < -n
\end{cases} = max(-n, d)
\]

$ max() $ is well-defined for intervals.
Specifically, for two intervals $ [a,b], [c,d] $, we have:
\[
	max([a,b], [c,d]) = [max(a, c), max(b, d)].
\]
Now plugging in the the bounds of $ n $ and $ d $ we get:
\begin{align*}
\deq(n,d) & = max\Big( [-\UBconcrete{\ReluIN{\n}}, -\LBconcrete{\ReluIN{\n}}], \\
& \qquad \quad \;\; [\LBconcrete{\ReluINdelta{\n}}, \UBconcrete{\ReluINdelta{\n}}]\Big)\\
 & = [max\Big(-\UBconcrete{\ReluIN{\n}}, \LBconcrete{\ReluINdelta{\n}} \Big),  \\
& \qquad \quad \;\; max\Big(-\LBconcrete{\ReluIN{\n}}, \UBconcrete{\ReluINdelta{\n}}\Big)]
\end{align*}

\ignore{To illustrate, let $ \ReluIN{\n} = [1,3] $ and $ \ReluINdelta{\n} = [-4,-2] $. 
Using the above property, the minimum difference is
$ \ReLU{3 + (-4)} - \ReLU{3} = 0 - 3 = -3 $, so we can tighten the 
lower bound to $ -3 $. The maximum difference is
$ \ReLU{1 + (-2)} - \ReLU{1} = 0 - 1 = -1 $, so we must increase the upper
bound.}

\subsubsection{Opt. 2: $ \n $ is non-linear, $ \np $ is active.}
Using Equation~\ref{eq:2}, we can tighten the upper bound.
We first rewrite Equation~\ref{eq:2} as:
\begin{align}
\deq'(n', d) = n' - \ReLU{n' - d}. \label{eq:5}
\end{align}
Just like Equation~\ref{eq:4}, Equation~\ref{eq:5} can be broken into
two cases based on the inequality $ n' - d \geq 0 = n' \geq d $, which 
gives us the piece-wise equation:
\[
\deq'(n', d) = \begin{cases}
d & n' \geq d\\
n' & n' < d
\end{cases} = min(n', d)
\]
For two intervals $[a,b],[c,d]$, we have
\[
min([a,b],[c,d]) = [min(a, c), min(b, d)].
\]
Replacing $ n' $ and $ d $ with the proper bounds gives us the
$ min() $ function in Algorithm~\ref{alg:relu}.

\ignore{To illustrate, let $ \ReluIN{\np} = [1, 3] $ and $ \ReluINdelta{\n} = [2, 4] $.
The minimum value is $ \ReLU{1} - \ReLU{1 - 2} = 1 $, and the
max value is $ \ReLU{3} - \ReLU{3-4} = 3 $, so we update the bounds
accordingly.}

\subsubsection{Opt. 3: both $ \n $ and $ \np $ are non-linear.} 
We consider three cases.
First, let $ \LBconcrete{\ReluINdelta{\n}} \geq 0 $.
This means that $ \np \geq \n $ before applying $ \proc{ReLU} $, and
then we can derive 0 as a lower bound as follows:
\begin{align*}
\np \geq \n \implies & \R{\np} \geq \R{\n} \\
&= \R{\np} - \R{\n} \geq 0
\end{align*}
In addition, $ \UBconcrete{\ReluINdelta{\n}} $
can be derived as an upper bound\footnote{
	In fact, the tighter upper bound $ min(\UBconcrete{\ReluINdelta{\n}}, \UBconcrete{\ReluIN{\np}}) $ can
	be derived, however we had not yet proved this at the time of submission.\label{fn:1}}
from Equation~\ref{eq:2}~\cite{arxivPaper}. 
This is the case in our motivating
example, so $ \ReluOUTdelta{n_{1,1}} = [0, 1.1] $. Second, we consider
$ \UBconcrete{\ReluINdelta{\n}} \leq 0 $. This means $\np \leq \n$ before $ \proc{ReLU} $, which allows
us to derive an upper bound of 0 in a symmetric manner to the first case. The lower bound shown in Algorithm~\ref{alg:relu}
can be derived from Equation~\ref{eq:1}~\cite{arxivPaper}. In the third case where
$ \LBconcrete{\ReluINdelta{\n}} < 0 < \UBconcrete{\ReluINdelta{\n}} $, the lower bound
and the upper bound shown in Algorithm~\ref{alg:relu} can be derived from Equations~\ref{eq:1} and~\ref{eq:2}, respectively~\cite{arxivPaper} (also see Footnote~\ref{fn:1}).

\subsection{On the Correctness}
The operations of the affine transformation are soundly defined for intervals as
described in Section~\ref{sec:int_analysis}. For the $ \proc{ReLU} $
transformation, we give formal explanations to show that they over-approximate (see also~\cite{arxivPaper} for proofs).
Since composing over-approximations also results in an over-approximation,
the forward analysis is itself a sound over-approximation.

\section{Gradient Based Refinement}
\label{sec:backward-pass}

After performing the forward pass, the computed difference may not be
tight enough to prove the desired property. In this section, we
discuss how we can improve the analysis result.

\subsection{Splitting Input Intervals}

As mentioned in Section~\ref{sec:prelims}, a common way to improve the
result of interval analysis is dividing an input interval into
disjoint sub-intervals, and then performing interval analysis on the
sub-intervals. 
After unioning the output intervals, the result will be \textit{at
least} as good as the original result~\cite{moore2009introduction}.
Prior work~\cite{WangPWYJ18} also shows a nice property of ReLU
networks: after a finite number of such splits, the result of the interval
analysis can be arbitrarily accurate.

However, determining the optimal order of refinement  is difficult and,
so far, the best algorithms are all heuristic based. For example, the
method used in \ReluVal{} chooses to split the input interval that has
\emph{the most influence} on the output value.
The intuition is that splitting such an input interval reduces the
approximation error of the output interval.

However, the approach is not suitable in our case because we focus on
the difference between the two networks: the input interval with the most
influence on the absolute value of the output may not have the most
influence on the output difference.  To account for this difference,
we develop a method for determining which input interval to split.

\subsection{The Refinement Algorithm}

Our idea is to compute the difference of the gradients for the two
networks, denoted $\nabla^\delta$.  Toward this end, we compute the
gradient of the first network ($\nabla$) and the gradient of the
second network ($\nabla'$).  Then, we use them to compute the
difference $\nabla^\delta$.

Formally, $\nabla = \{ \partial f/\partial x_1, \ldots, \partial
f/\partial x_n\}$ is a vector whose $i$-th element, $\nabla_i
= \partial f/\partial x_i$, is the partial derivative of the output
$f$ with respect to the input $x_i$.  Similarly, $\nabla' = \{\partial
f'/\partial x_1,\dots, \partial f'/\partial x_n\}$.
The difference is $\nabla^\delta = \{\partial (f'-f)/\partial
x_1, \dots, \partial (f'-f)/\partial x_n\}$ = $\nabla' - \nabla$.
That is, the derivative of a difference of functions is the difference
of their derivatives.

During interval analysis, the accurate gradient is difficult to
compute. Therefore, we compute the approximated gradient, where each
element $\nabla_i$ is represented by a concrete interval.

\begin{algorithm}[t]
\caption{Computing the gradient of a network.}
\label{alg:gradient}
{\footnotesize
	\SetAlgoLined
	\KwIn{network $f$,  mask matrix $R$}
	\KwResult{gradient $\nabla$}
        \textcolor{dkgreen}{// Initialize to edge weights in the output layer}\\
	$ \UB{\nabla_1} = \LB{\nabla_1} = 1 $\\
	\For{$ k = l-1..2 $}{
		$\nabla^{New} = \{ [0,0], \dots, [0,0] \}$\\
		\For{$ j = 1..layerSize[k] $}{
			\textcolor{dkgreen}{// Perform ReLU for node $n_{k,j}$}\\
			\uIf{$ R[k][j] == [0, 0] $} {
				$ \LB{\nabla_j} = \UB{\nabla_j} = 0 $
			} \uElseIf {$ R[k][j] = [0,1] $} {
				$ \LB{\nabla_j} = min(0, \LB{\nabla_j}) $\\
				$ \UB{ \nabla_j} = max(0, \UB{ \nabla_j}) $		
			}
			\textcolor{dkgreen}{// Multiply by weights of incoming edges to node $n_{k,j}$}\\
			\For{$ i = 1..layerSize[k-1] $}{
				\uIf{$ W_{k}[i,j] \geq 0$}{
					$ \UB{\nabla^{New}_i}  \pluseq W_k[i,j]*\UB{ \nabla_j} $\\
					$ \LB{\nabla^{New}_i} \pluseq W_k[i,j]*\LB{\nabla_j} $\\
				} \uElse {
					$ \UB{\nabla^{New}_i}  \pluseq W_k[i,j]*\LB{\nabla_j} $\\
					$ \LB{\nabla^{New}_i} \pluseq W_k[i,j]*\UB{ \nabla_j} $\\
				}
			}
		}
		$ \nabla = \nabla^{NEW}$ 
	}
	\Return{$ \nabla $}
}
\end{algorithm}

Algorithm~\ref{alg:gradient} shows our gradient computation procedure.
In addition to the network, which may be either $f$ or $f'$, it also
takes the mask matrix $R$ as input.  Recall that both $R[k][j]$ and
$R'[k][j]$ have been computed by Algorithm~\ref{alg:relu} during the
forward pass.  $R[k][i]$ may be $[0,0]$, $[1,1]$, or $[0,1]$,
indicating if the ReLU in $n_{k,j}$ is \emph{always
de-activated}, \emph{always activated}, or \emph{non-linear},
respectively.  It can be understood as the gradient interval of the
ReLU.

The gradient computation is performed backwardly beginning at the
output layer and then moving through the previous layers. In each
layer, the computation has two steps. First we apply ReLU to the
current gradient and update the upper and lower bounds of the gradient
if needed. Then, we scale the gradient interval by the weights of the previous
layer.

After computing $\nabla$ and $\nabla'$ by invoking
Algorithm~\ref{alg:gradient} on $f$ and $f'$, respectively, we compute
the gradient difference $\nabla^\delta$.

Then, we use the gradient difference to determine which input has the
most influence on the output difference.  Note that the gradient
itself is not sufficient to act as an indicator of influence.  For
example, while an input's gradient may be large, but the width of
its input interval is small, splitting it will not have much impact on
the output interval.
Thus, we split the input interval with the maximum \textit{smear}
value~\cite{kearfott1990algorithm, kearfott2013rigorous}.  The smear value of an input $x_i$ is defined as the width
of its input interval $|\UBconcrete{x_i} - \LBconcrete{x_i}|$ scaled
by the upper bound of its corresponding gradient difference
$\UBconcrete{\nabla^\delta}_i$.

\subsection{An Example}

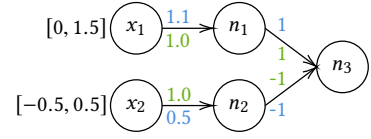
\begin{figure}
	\scalebox{0.65}{
	\tikzset{every picture/.style={line width=0.75pt}} %set default line width to 0.75pt        

\begin{tikzpicture}[x=0.75pt,y=0.75pt,yscale=-1,xscale=1]
%uncomment if require: \path (0,300); %set diagram left start at 0, and has height of 300

%Shape: Circle [id:dp06612404457840415] 
\draw   (80,60) .. controls (80,48.95) and (88.95,40) .. (100,40) .. controls (111.05,40) and (120,48.95) .. (120,60) .. controls (120,71.05) and (111.05,80) .. (100,80) .. controls (88.95,80) and (80,71.05) .. (80,60) -- cycle ;
%Shape: Circle [id:dp685015234974003] 
\draw   (80,120) .. controls (80,108.95) and (88.95,100) .. (100,100) .. controls (111.05,100) and (120,108.95) .. (120,120) .. controls (120,131.05) and (111.05,140) .. (100,140) .. controls (88.95,140) and (80,131.05) .. (80,120) -- cycle ;
%Shape: Circle [id:dp36531807362590973] 
\draw   (160,60) .. controls (160,48.95) and (168.95,40) .. (180,40) .. controls (191.05,40) and (200,48.95) .. (200,60) .. controls (200,71.05) and (191.05,80) .. (180,80) .. controls (168.95,80) and (160,71.05) .. (160,60) -- cycle ;
%Shape: Circle [id:dp8605838523620923] 
\draw   (160,120) .. controls (160,108.95) and (168.95,100) .. (180,100) .. controls (191.05,100) and (200,108.95) .. (200,120) .. controls (200,131.05) and (191.05,140) .. (180,140) .. controls (168.95,140) and (160,131.05) .. (160,120) -- cycle ;
%Shape: Circle [id:dp18951934248445623] 
\draw   (240,90) .. controls (240,78.95) and (248.95,70) .. (260,70) .. controls (271.05,70) and (280,78.95) .. (280,90) .. controls (280,101.05) and (271.05,110) .. (260,110) .. controls (248.95,110) and (240,101.05) .. (240,90) -- cycle ;
%Straight Lines [id:da2027373665168457] 
\draw    (120,60) -- (158,60) ;
\draw [shift={(160,60)}, rotate = 180] [color={rgb, 255:red, 0; green, 0; blue, 0 }  ][line width=0.75]    (10.93,-3.29) .. controls (6.95,-1.4) and (3.31,-0.3) .. (0,0) .. controls (3.31,0.3) and (6.95,1.4) .. (10.93,3.29)   ;

%Straight Lines [id:da534577266557194] 
\draw    (120,120) -- (158,120) ;
\draw [shift={(160,120)}, rotate = 180] [color={rgb, 255:red, 0; green, 0; blue, 0 }  ][line width=0.75]    (10.93,-3.29) .. controls (6.95,-1.4) and (3.31,-0.3) .. (0,0) .. controls (3.31,0.3) and (6.95,1.4) .. (10.93,3.29)   ;

%Straight Lines [id:da6712362717076668] 
\draw    (200,120) -- (238.4,91.2) ;
\draw [shift={(240,90)}, rotate = 503.13] [color={rgb, 255:red, 0; green, 0; blue, 0 }  ][line width=0.75]    (10.93,-3.29) .. controls (6.95,-1.4) and (3.31,-0.3) .. (0,0) .. controls (3.31,0.3) and (6.95,1.4) .. (10.93,3.29)   ;

%Straight Lines [id:da7166627548154052] 
\draw    (200,60) -- (238.4,88.8) ;
\draw [shift={(240,90)}, rotate = 216.87] [color={rgb, 255:red, 0; green, 0; blue, 0 }  ][line width=0.75]    (10.93,-3.29) .. controls (6.95,-1.4) and (3.31,-0.3) .. (0,0) .. controls (3.31,0.3) and (6.95,1.4) .. (10.93,3.29)   ;

% Text Node
\draw (132.5,51) node [scale=1.44,color={rgb, 255:red, 74; green, 144; blue, 226 }  ,opacity=1 ] [align=left] {1.1};
% Text Node
\draw (133,129) node [scale=1.44,color={rgb, 255:red, 74; green, 144; blue, 226 }  ,opacity=1 ] [align=left] {0.5};
% Text Node
\draw (213,57) node [scale=1.44,color={rgb, 255:red, 74; green, 144; blue, 226 }  ,opacity=1 ] [align=left] {1};
% Text Node
\draw (210,99) node [scale=1.44,color={rgb, 255:red, 105; green, 179; blue, 20 }  ,opacity=1 ] [align=left] {\mbox{-}1};
% Text Node
\draw (100,60) node [scale=1.44]  {$x_1$};
% Text Node
\draw (100,120) node [scale=1.44]  {$x_2$};
% Text Node
\draw (53.5,60) node [scale=1.44]  {$[ 0,1.5]$};
% Text Node
\draw (42,120) node [scale=1.44]  {$[ -0.5,0.5]$};
% Text Node
\draw (180,60) node [scale=1.44]  {$n_{1}$};
% Text Node
\draw (180,120) node [scale=1.44]  {$n_{2}$};
% Text Node
\draw (260,90) node [scale=1.44]  {$n_{3}$};
% Text Node
\draw (132,69) node [scale=1.44,color={rgb, 255:red, 105; green, 179; blue, 20 }  ,opacity=1 ] [align=left] {1.0};
% Text Node
\draw (133,111) node [scale=1.44,color={rgb, 255:red, 105; green, 179; blue, 20 }  ,opacity=1 ] [align=left] {1.0};
% Text Node
\draw (213,79) node [scale=1.44,color={rgb, 255:red, 105; green, 179; blue, 20 }  ,opacity=1 ] [align=left] {1};
% Text Node
\draw (210,123) node [scale=1.44,color={rgb, 255:red, 74; green, 144; blue, 226 }  ,opacity=1 ] [align=left] {\mbox{-}1};

\end{tikzpicture}	
	}
	\caption{Example for backward refinement.\label{fig:refine_ex}}
\end{figure}

We now walk through the gradient computation in
Algorithm~\ref{alg:gradient} for the example in
Figure~\ref{fig:refine_ex}, where blue weights are for network $f$,
and green weights are for network $f'$.  We focus on the gradient of
$f$ first.  After performing the forward pass, we know that $n_1$ is
in a linear state, i.e., $ R[1][0] = [1,1] $, and $ n_2 $ is in a
non-linear state, i.e., $ R[1][1] = [0,1] $.

We initialize the gradient to the weights of the final layer; that is,
$ \UB{\nabla_1} = \LB{\nabla_1} = 1 $ and $ \UB{\nabla_2}
= \LB{\nabla_2} = -1 $.
Next, we apply ReLU. Since $ n_1 $ is in the \emph{always activated}
mode, we leave its gradient unchanged. However, $ n_2 $ is in
the \emph{non-linear} mode, meaning the gradient could be 0, and hence
we must ensure that 0 is in the gradient interval.  We update
$ \LB{\nabla_2} = 0 $.
Then, we scale the gradient interval by weights of the incoming edges,
which gives us the gradient intervals for input variables: $ \nabla_1
= [1.1, 1.1] $ for $x_1$ and $ \nabla_2 = [-0.5, 0] $ for $x_2$.

Here, we point out a problem with \ReluVal{}'s refinement method.  It
would compute the smear value of $ x_1 $ and $ x_2 $ to be $ |(1.5
- 0)*1.1| = 1.65 $ and $ |(0.5 - (-0.5))*-0.5| = 0.5 $, respectively,
which means it would split on $ x_1 $.  However, this is not appropriate for
differential verification, since the two networks differ the most in
the weights of the outgoing edge of $x_2$.

Our method, instead, would compute the gradient difference
$\nabla^\delta = \nabla - \nabla'$.  Therefore, we have
$\nabla^\delta_1 = [-0.1, -0.1]$ for $ x_1 $ and $ \nabla^\delta_2 =
[-0.5, 1] $ for $ x_2 $.  Based on the new smear values, we would
choose to split the input interval of $x_2$.

%\clearpage
%\newpage
\section{Experiments}
\label{sec:experiment}

We have implemented \diffNN{} and compared it experimentally with
state-of-the-art neural network verification tools.  Like \ReluVal,
\diffNN{} is written in C using OpenBLAS~\cite{ZhangWZ12} as the
library for matrix multiplications. We also note that we implement outward-rounding to soundly handle floating point arithmetic. Symbolic interval arithmetic is
implemented using matrix multiplication.  \diffNN{} takes two
networks $f$ and $f'$ together with a small $\epsilon$ and input
region $ X $ as input, and then decides whether $\forall
x \in X$. $|f'(x)-f(x)|<\epsilon$ for the target label's value.
Since \diffNN{} is the only tool currently available for differential
verification of neural networks, to facilitate the experimental
comparison with existing tools, we developed a tool to
merge $f$ and $f'$ into a combined network $f''$, as shown in
Figure~\ref{fig:composed-network}, before feeding $f''$ to these existing
tools as input.

\subsection{Benchmarks}

Our benchmarks are 49 feed-forward neural networks
from three applications: aircraft collision detection, image
recognition, and human activity recognition. We produce $ f' $ by
truncating each network's weights from 32-bit floats to 16-bit
floats.

\subsubsection{ACAS Xu~\cite{JulianKO18}}

ACAS Xu is a set of 45 neural networks commonly used in evaluating
neural network verification tools. They are designed to be used in an
aircraft to advise the pilot of what action to take in the presence of
an intruder aircraft. They each take five inputs: distance between
self and the intruder, angle of self relative to the intruder, angle
of intruder relative to self, speed of self, and speed of
intruder. They output a score in the range $[-0.5, 0.5]$ for five
different actions: clear-of-conflict, weak left, weak right, strong
left, and strong right. The action with the minimum score is the action
advised.  In addition to the input and output layers, each network has
6 hidden layers of 50 neurons each, for a total of 300 neurons.  For
differential verification, we use the same input ranges as
in \cite{KatzBDJK17, WangPWYJ18} as our regions of interest.

\subsubsection{MNIST~\cite{LecunBBH98}}

MNIST is a data set of labeled images of hand-written digits that are
often used as a benchmark to test image classifiers. The images are
28x28 pixels, and each pixel has a gray-scale value in the range $[0,
255]$. Neural networks trained on this data set take in 784 inputs
(one per pixel) each in the range $[0, 255]$, and output 10
scores, typically in the range of $[-10, 10]$ for our networks, for each of the 10
digits. The digit with the highest score is the chosen classification.
We use three neural networks trained on the MNIST data set with
architectures of 3x100, 2x512, and 4x1024; that is, the networks have
3, 2, and 4 hidden layers, with layer size 100, 512, and 1024 neurons, respectively.  Thus,
in addition to the input and output layers, these networks have 
300, 1024, and 4096 hidden neurons, respectively.  Empirical
analysis shows that each network has > 95\% accuracy on hold-out test
data.

\subsubsection{Human Activity Recognition (HAR)~\cite{AnguitaHAR}}
HAR is a labeled data set used to train models to recognize specific
human activities based on input from a smartphone's accelerometer and gyroscope.
Input examples in this data set are labeled with one of six activities: 
walking, walking upstairs, walking downstairs, sitting,
standing, and laying down. The input data for the model are statistics computed from a smartphone's accelerometer and gyroscope sensor, 
such as mean, median, min, max, etc. In total, 561 input statistics are computed from these two
sensors. Inputs to the network are normalized to be in the range of $[-1, 1]$.
We use a network trained on this data set with an architecture of
1x500, meaning there is a hidden layer with 500 neurons. The network
takes the 561 inputs, and produces a score in the range
of $[-20, 20]$ for each of the 6 outputs, one per
activity.  The output with the maximum score is the classification.

Table~\ref{tbl:benchmark} shows the statistics of these benchmarks, including the number
of input neurons, the number of output neurons, the number of hidden
layers, as well as the total number of neurons in these hidden layers. The last two columns list the experimental parameters we used, namely the number of ``regions of interest'' in the verification problems and the output $ \epsilon $ we attempt to verify.

\begin{table}
\caption{Statistics of the benchmarks: The total number of verification problems is 842.}
\label{tbl:benchmark}
\scalebox{0.73}{
\begin{tabular}{|l|c|c|c|c|c|c|c|}\hline
Name                      & \#  & \multicolumn{4}{c|}{in  each network}             & \# input    & out \\
\cline{3-6}
                          &   NN's      &  \# in & \# out & \# hidden  & \# neurons      & region  & $\epsilon$ \\
\hline\hline
ACAS-$\phi_1$-$\phi_2$ &   45    &   5    &     5       &      6 * 50   &    300     & 1           & 0.05  \\
ACAS-$\phi_3$          &   42    &   5    &     5       &      6 * 50   &    300     & 1           & 0.05 \\
ACAS-$\phi_4$          &   42    &   5    &     5       &      6 * 50   &    300     & 1           & 0.05 \\
ACAS-$\phi_5$-$\phi_{13} $ &   1 &   5    &     5       &      6 * 50   &    300     & 1  		   & 0.05 \\
ACAS-$\phi_{14}$       &   2    &   5    &     5       &      6 * 50   &    300     & 1           & 0.05 \\
ACAS-$\phi_{15}$       &   2    &   5    &     5       &      6 * 50   &    300     & 1           & 0.05  \\
\hline
MNIST 3x100     &    1    & 784    &     10      &      3 * 100  &    300     & 200           & 1 \\
MNIST 2x512     &    1    & 784    &     10      &      2 * 512  &  1,024     & 200           & 1 \\
MNIST 4x1024    &    1    & 784    &     10      &      4 * 1024 &  4,096     & 200           & 1 \\
\hline
HAR   1x500     &    1    & 561    &     6       &      1 * 500  &    500     & 100          &   0.25 \\
\hline
\end{tabular}
}
\end{table}

\subsection{Experimental Evaluation}

We want to answer the following research questions:
\begin{enumerate}
\item Is \diffNN{} more efficient than existing methods in differential verification of neural networks 
		in that it can both verify properties faster and verify more properties in general?
\item Is \diffNN{} more accurate than existing methods in the forward pass?
\end{enumerate}
Toward this end, we directly compared \diffNN{} to two
state-of-the-art verification tools: \ReluVal{}~\cite{WangPWYJ18}
and \DeepPoly{}~\cite{SinghGPV19}. Both are designed to
formally verify the absence of adversarial examples. 
\ignore{Since \diffNN{}
and \ReluVal{} accept the same type of networks as input, we are able
to compare them on all 292 verification problems.  \DeepPoly{}, in
contrast, does not support the input range format for our ACAS Xu and
HAR benchmarks, so we are not able to have a direct comparison;
nevertheless, we are able to compare with \DeepPoly{} on the
MNIST benchmarks.}

A comparison with \Reluplex{}~\cite{KatzBDJK17} was not possible since
it does not support affine hidden layers, which are necessary for
analyzing the combined network $f''(x)$ as shown in Figure~\ref{fig:composed-network}, however we note
that \ReluVal{} previously has been shown to significantly outperform \Reluplex{} on
all ACAS Xu benchmarks~\cite{WangPWYJ18}.
\DeepPoly{},  a followup of \textsc{AI2}~\cite{GehrMDTCV18}, has also 
been shown to outperform \textsc{AI2}.
\ignore{We also attempted to run \RefineZono{}~\cite{SinghGPV19iclr}, however
the tool was crashing on our benchmarks and the authors were actively
fixing the tool at the time of writing.}

We ran all experiments on a Linux server running Ubuntu 16.04, an Intel Xeon CPU E5-2620, and 124 GB memory. Timeout is set at 30 minutes for each verification
problem.  When available, we enable parallelization in all tools and
configure them to allow up to 10 threads at a time.

\subsection{Results}

To evaluate efficiency and accuracy, we truncate each network's weights from 32-bit floats to 16-bit floats, and attempt to verify the $ \epsilon $ shown in Table~\ref{tbl:benchmark}. We measure the number of properties verified and the execution time to verify each property.

\begin{table}
%
% for a table, the caption should be at the top
% for a figure, the caption should be at the bottom
%
%\caption{Accuracy comparison of \diffNN{} and \ReluVal{} on ACAS networks, where network $f'$ is network $f$ rounded to 2 significant figures, and $\epsilon=0.1$. }
\caption{Accuracy of comparison of the three tools on ACAS.}
\label{tbl:ACAS_accuracy}
\scalebox{0.75}{
\begin{tabular}{|l|c|cc|cc|cc|}\hline
Benchmark              & Verif.       & \multicolumn{2}{c|}{\diffNN{} (new)} & \multicolumn{2}{c|}{\ReluVal{}} & \multicolumn{2}{c|}{\DeepPoly} \\
\cline{3-8}
                       & problems     & proved  & undet. & proved & undet. & proved & undet. \\\hline
\hline
ACAS $\phi_1$-$\phi_2$  & 45	&   28    &  17     &    7   &   38     &  0  &   45              \\\hline
ACAS $\phi_3$           & 42	&   42    &   0     &   24   &   18     &  6  &   36              \\\hline
ACAS $\phi_4$	        & 42	&	42    &   0     &   34   &   8     &   2  &   40              \\\hline
ACAS $\phi_5$           &  1	&	1    &   0     &    0   &    1     &   0  &   1           \\\hline
ACAS $\phi_6$           &  1	&	1    &   0     &    1   &    0     &   0  &   1     \\\hline
ACAS $\phi_7$           &  1	&	0    &   1     &    0   &    1     &   0  &   1      \\\hline
ACAS $\phi_8$           &  1	&	0    &   1     &    0   &    1     &   0  &   1      \\\hline
ACAS $\phi_9$           &  1	&	1    &   0     &    0   &    1     &   0  &   1       \\\hline
ACAS $\phi_{10}$        &  1	&	1    &   0     &    1   &    0     &   0  &   1       \\\hline
ACAS $\phi_{11}$        &  1	&	1    &   0     &    0   &    1     &   0  &   1      \\\hline
ACAS $\phi_{12}$        &  1	&	0    &   1     &    0   &    1     &   0  &   1      \\\hline
ACAS $\phi_{13}$        &  1	&	1    &   0     &    1   &    0     &   0  &   1      \\\hline
ACAS $\phi_{14}$        &  2	&	2    &   0     &    0   &    2     &   0  &   2      \\\hline
ACAS $\phi_{15}$        &  2	&	2    &   0     &    0   &    2     &   0  &   2      \\\hline
\hline
Total                   & 142         &    123   &  19     &   69   &    73    & 8  & 134 \\\hline
\end{tabular}
}
\end{table}

\begin{table}
%
% for a table, the caption should be at the top
% for a figure, the caption should be at the bottom
%
%\caption{Efficiency comparison of \diffNN{} and \ReluVal{} on ACAS networks, where network $f'$ is network $f$ rounded to 2 significant figures, and $\epsilon=0.1$. }
\caption{Efficiency of \diffNN{} vs. \ReluVal{} on ACAS Xu. }
\label{tbl:ACAS_efficiency}
\scalebox{0.75}{
\begin{tabular}{|l|c|r|r|r|}\hline
Benchmark         & Verif.      & \multicolumn{3}{c|}{ Total Time (s) }\\\cline{3-5}
                  & problems    & \diffNN{} (new)     & \ReluVal{}        &  Avg. Speedup \\\hline
\hline
ACAS $\phi_{1--2}$ 	&	45	&	40595.6	&	69167.5		&	$ 1.7 $			\\\hline
ACAS $\phi_{3}$    	&	42	&	175.4	&	38414.2		&	$ \geq 219 $	\\\hline
ACAS $\phi_{4}$   	&	42	&	46.8	&	22159.2		&	$ \geq 473.4 $	\\\hline
ACAS $\phi_{5}$   	&   1	&	9.6		&	1800.0		&	$ \geq 187.5 $	\\\hline
ACAS $\phi_{6}$   	&   1	&	11.0	&	50.8		&	$ 4.6 $			\\\hline
ACAS $\phi_{7}$   	&   1	&	1800.0	&	1800.0		&	$ 1.00$			\\\hline
ACAS $\phi_{8}$   	&   1	&	1800.0	&	1800.0		&	$ 1.00$			\\\hline
ACAS $\phi_{9}$   	&   1	&	52.3	&	1800.0		&	$ \geq 34.4 $	\\\hline
ACAS $\phi_{10}$  	&   1	&	31.0	&	53.3		&	$ 1.6 $			\\\hline
ACAS $\phi_{11}$  	&   1	&	10.2	&	1800.0		&	$ \geq 177.3 $	\\\hline
ACAS $\phi_{12}$  	&   1	&	1800.0	&	1800.0		&	$ 1.0 $			\\\hline
ACAS $\phi_{13}$  	&   1	&	157.9	&	999.2		&	$ 6.3 $			\\\hline
ACAS $\phi_{14}$  	&   2	&	859.0	&	3600.0		&	$ \geq 4.2 $	\\\hline
ACAS $\phi_{15}$  	&   2	&	453.8	&	3600.0		&	$ \geq 7.9  $	\\\hline
\end{tabular}
}
\end{table}

\subsubsection{ACAS Xu} 
The results for ACAS Xu are shown in Tables~\ref{tbl:ACAS_accuracy} and~\ref{tbl:ACAS_efficiency}. 
In Table~\ref{tbl:ACAS_accuracy}, columns 1 and 2 show the input property
used, and the number of networks we verified the property on, which are taken from~\cite{KatzBDJK17, WangPWYJ18}. 
Columns 3-5 show the number of neural networks for which the
property was verified and undetermined for each tool. 
Undetermined means that either the tool reported it could not verify the problem (due to over-approximation), 
or the timeout of 30 minutes was reached.

In Table~\ref{tbl:ACAS_efficiency}, Columns 3-4 show the time taken
by \diffNN{} and \ReluVal{} for all problems verified. The last column
shows the time of \ReluVal{} divided by the time of \diffNN{}.  For
timeouts, we add 30 minutes to the total, which is why we display that the
speedup is \textit{greater than or equal to }$ X $ for some
properties. We omit the timing data for \DeepPoly{} since it cannot
verify most properties.

These results emphasize the improvement that \diffNN{} can obtain in both
speed and accuracy. It achieves orders of magnitude speedups over \ReluVal. For
example, \diffNN~finishes the 42 networks for $ \phi_4 $ in 48.6
seconds, whereas \ReluVal~ takes \textit{at least} more than 6
hours. Overall \diffNN~ verifies 54 more problems for
which \ReluVal~times out, and 115 more problems for which \DeepPoly{}
is too inaccurate to verify.

\begin{figure}[t]
	\begin{minipage}[t]{0.49\linewidth}
		\captionsetup{width=0.8\textwidth}
		\includegraphics[width=\linewidth]{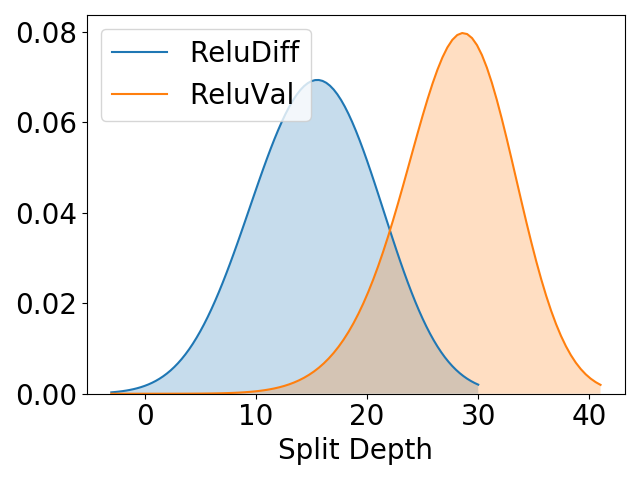}
		\caption{$ \phi_4 $ max depth distribution.\label{fig:prop4_depth}}
	\end{minipage}
	\begin{minipage}[t]{0.495\linewidth}
				\captionsetup{width=0.9\textwidth}
			\includegraphics[width=\linewidth]{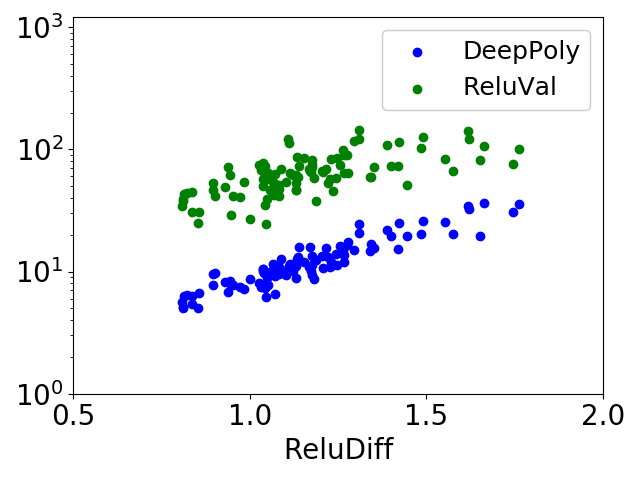}
			\caption{$\Delta $-interval on MNIST 4x1024\label{fig:deeppoly_comp}.}
	\end{minipage}%
\end{figure}

To understand why \diffNN{} performs better, we plot the distribution
of the depth at which each sub-interval was finally able to be
verified for $ \phi_4 $, shown in Figure~\ref{fig:prop4_depth}. We
can see that \diffNN{} consistently verifies sub-intervals at much
shallower split depths. We point out that the number of sub problems grows
exponentially as the split depth increases. Indeed, even though the
difference between the average depths does not seem large (about 14
for \diffNN{} and 29 for \ReluVal{}), \ReluVal{} had to verify > 66 million
sub-intervals for $ \phi_4 $, whereas \diffNN{} only had to verify
66K.

\subsubsection{MNIST} 

While in ACAS Xu the input region to verify is defined by the
property, for MNIST, we must generate the input region ourselves. We
generate 200 input regions for MNIST using two methods. The first
method is based on global perturbation~\cite{SinghGPV19}. We take 100
test images, and for each one, we allow each of the pixels to be
perturbed by +/-3 gray scale units. The second method is based on
targeted pixel perturbation~\cite{GopinathPWZK19,GopinathKPB18}. We
take the same 100 test images, and for each one, we set the range of 3 random
pixels to $ [0,255] $, while the remaining 781 remain fixed.

\begin{table}
%\caption{Accuracy comparison of \diffNN{}, \ReluVal{} and \DeepPoly{} on MNIST networks, where network $f'$ is derived from network $f$ by rounding the weights to 3 significant figures, and setting  $\epsilon=2.5$.}
\caption{Accuracy comparison of the three tools on MNIST.}
\label{tbl:MNIST_accuracy}
\scalebox{0.75}{
\begin{tabular}{|l|c|cc|cc|cc|}\hline
Benchmark & Verif. & \multicolumn{2}{c|}{\diffNN{} (new)}
& \multicolumn{2}{c|}{\ReluVal{}} & \multicolumn{2}{c|}{\DeepPoly} \\
\cline{3-8}
                & problems & proved  & undet. & proved & undet. & proved & undet. \\\hline
\hline

3x100-global     & 100      & 100     &   0    &   47    & 53     & 34    & 66     \\\hline
2x512-global     & 100      & 100     &   0    &   0    & 100     & 0    & 100     \\\hline
4x1024-global    & 100      & 22     &  78    &   0    & 100     & 0   & 100      \\\hline
\hline
3x100-3-pixel     & 100     & 100     &   0    &   100    & 0     & 100    & 0     \\\hline
2x512-3-pixel     & 100     & 100     &   0    &   100    & 0     & 80   & 20     \\\hline
4x1024-3-pixel    & 100     & 100     &  0    &   97    & 3     & 100   & 0      \\\hline

\end{tabular}
}
\end{table}

\begin{table}
%\caption{Efficiency comparison of \diffNN{}, \ReluVal{} and \DeepPoly{} on MNIST networks, where network $f'$ is derived from network $f$ by rounding the weights to 3 significant figures, and setting  $\epsilon=2.5$.}
\caption{Efficiency comparison of the three tools on MNIST.}
\label{tbl:MNIST_efficiency}
\scalebox{0.75}{
\begin{tabular}{|l|c|r|r|r|}\hline
Benchmark         & Verif.  & \multicolumn{3}{c|}{ Total Time (s) }\\\cline{3-5}
                  & problems & \diffNN{} (new) & \ReluVal{} & \DeepPoly{} \\\hline
\hline
3x100-global       &     100    &        29.47     &        95458.32          &   118823.09          \\\hline
2x512-global       &     100    &        77.83     &        180000.00          &   180000.0          \\\hline
4x1024-global      &     100    &     141604.53    &        180000.00       &      180000.0     \\\hline
\hline
3x100-3-pixel       &     100   &     23.90    &        32.60            &   163.75         \\\hline
2x512-3-pixel       &     100   &     79.24    &        715.16             &   37674.40         \\\hline
4x1024-3-pixel      &     100   &     296.59       &     92100.10          &    49042.98    \\\hline
\end{tabular}
}
\end{table}

We can again see in Tables~\ref{tbl:MNIST_accuracy}
and \ref{tbl:MNIST_efficiency} that \diffNN{} is significantly more
accurate and efficient than both \ReluVal{} and \DeepPoly{}.
Both competing techniques struggle to handle global perturbations even
on the small 3x100 network, let alone the larger 2x512 and 4x1024 networks.
On the other hand, \diffNN{} can easily handle both the 3x100 and 2x512 networks, 
achieving at least 3 orders of magnitude speedup on these networks. 
We also see a three orders of magnitude speedup on the two largest networks 
for our targeted-pixel perturbation experiments.

Even though \diffNN{} begins to reach its limit in the global perturbation experiment 
on the largest 4x1024 network, we point out that \diffNN{} is significantly outperforming
both \DeepPoly{} and \ReluVal{} in the accuracy of their forward passes.
Figure~\ref{fig:deeppoly_comp} compares the output bound verified on the
\textit{first, single} forward pass of each technique.
The comparison is presented as a scatter plot, where the x-axis is the
bound verified by \diffNN{}, and the y-axis is that of the competing technique.

The graph shows that \diffNN{} is nearly two orders of magnitude more accurate
than \ReluVal{} and one order of magnitude more than \DeepPoly.
The improvement over \DeepPoly{} especially emphasizes the promise of \diffNN{}'s
approach. This is because \diffNN{} is already outperforming \DeepPoly{}, yet it
uses a simpler \emph{concretization} approach during the forward pass, whereas
\DeepPoly{} uses a more sophisticated \emph{linear relaxation}.
We believe that \diffNN{} can be extended to use more accurate techniques such as 
linear relaxation which would further improve the accuracy, however we leave 
this as future work.

\subsubsection{HAR} 

For HAR, we also created our verification problems using input
perturbation.  We take 100 concrete test inputs, and for each one, we
allow a global perturbation of +/-0.1. 
The results are summarized in Tables~\ref{tbl:HAR_accuracy} and~\ref{tbl:HAR_efficiency}.
Again, the experimental comparison shows that \diffNN{} is
significantly more accurate and efficient.

\begin{table}
	\caption{Accuracy comparison of the three tools on HAR.}
	\label{tbl:HAR_accuracy}
	\scalebox{0.75}{
		\begin{tabular}{|l|c|cc|cc|cc|}\hline
			Benchmark & Verif. & \multicolumn{2}{c|}{\diffNN{} (new)}
			& \multicolumn{2}{c|}{\ReluVal{}} & \multicolumn{2}{c|}{\DeepPoly} \\
			\cline{3-8}
			& problems & proved  & undet. & proved & undet. & proved & undet. \\\hline
			\hline
			
			1x500     & 100      & 100     &   0    &   0    &   100   &  0 &  100   \\\hline
			
		\end{tabular}
	}
\end{table}

\begin{table}
	\caption{Efficiency comparison of the three tools on HAR.}
	\label{tbl:HAR_efficiency}
	\scalebox{0.75}{
		\begin{tabular}{|l|c|r|r|r|}\hline
			Benchmark         & Verif.  & \multicolumn{3}{c|}{ Total Time (s) }\\\cline{3-5}
			& problems & \diffNN{} (new) & \ReluVal{} & \DeepPoly{} \\\hline
			\hline
			1x500       &     100    &        28.79     &        180000.00   &   180000.00   \\\hline
		\end{tabular}
	}
\end{table}

\subsection{Threats to Validity}

Our method is designed for verifying neural networks typically found
in control applications, where the number of input signals is not
large. In this context, dividing the input region turns out to be a
very effective way of increasing the accuracy of interval analysis.
However, neural networks in different application domains may have
different characteristics.  Therefore, it remains an open problem
whether bi-section of individual input intervals is always an
effective way of performing refinement.

Our method is designed for feed-forward ReLU networks.  Although there
is no significant technical hurdle for it to be extended to
convolutional neural networks or other activation functions, such as
sigmoid, tanh and max-pool as shown recently by Singh et
al.~\cite{SinghGPV19}, we have not evaluated the effectiveness.
Specifically, linear relaxation can be used to handle these features
when it comes to approximating non-linear behavior.  While we use
concretization in \diffNN{}, extending it with linear relaxation is
possible~\cite{WangPWYJ18nips}. However, we leave these extensions for
future work.

\section{Related Work}
\label{sec:related}

While there is a large and growing body of work on detecting
adversarial examples for neural networks, they are typically based on
heuristic search or other dynamic analysis techniques such as
testing~\cite{CarliniW17,PeiCYJ17,TianPJR18,SunWRHKK18,WickerHK18,MaLLZG18}.
Although they are effective in finding security vulnerabilities and
violations of other critical properties, we consider them as being
orthogonal to formal verification.  The reason is because these
techniques are geared toward finding violations, as opposed to proving
the absence of violations.

Early work on formal verification of deep neural networks relies on
using SMT solvers~\cite{HuangKWW17,Ehlers17}, or SMT solving
algorithms~\cite{KatzBDJK17,KatzHIJLLSTWZDK19} designed for
efficiently reasoning about constraints from the ReLU activation
function. Along this line, a state-of-the-art tool
is \Reluplex{}~\cite{KatzBDJK17}.  In theory, these SMT solver based
techniques can solve the neural network verification problem in a
sound and complete fashion, i.e., returning a proof if and only if the
network satisfies the property.  In practice, however, their
scalability is often limited and they may run out of time for larger
networks.

Another line of work on verification of deep neural networks is based
on interval analysis, which can be more scalable than
SMT solver based techniques~\cite{WangPWYJ18}.  They compute
conservative bounds on the value ranges of the neurons and output
signals for an input region of interest.  They also exploit the fact
that neural networks are Lipschitz continuous~\cite{RuanHK18} to
ensure that the interval analysis results are
sound.  \ReluVal{}~\cite{WangPWYJ18} and \DeepPoly{}~\cite{SinghGPV19}
are two representatives, among other similar
tools~\cite{WangPWYJ18nips,SinghGPV19iclr,MirmanGV18,GehrMDTCV18,FischerBDGZV19}.

In addition to formal verification, there are techniques for
evaluating and certifying the robustness of neural
networks~\cite{BastaniILVNC16,CarliniW17,WengZCSHDBD18,DvijothamSGMK18}
or certified defense against adversarial
examples~\cite{RaghunathanSL18,WongK18}.
However, neither they nor the existing verification techniques
were designed for \emph{differential verification} of two closely
related neural networks, which is the focus of this paper. 
As shown by the examples in Section~\ref{sec:motivation} and the
experimental results in Section~\ref{sec:experiment}, directly
applying these techniques to differential verification is often
extremely inefficient.  In contrast, our method is designed
specifically for solving the differential verification problem efficiently.

At a higher level, our method relies on symbolic interval analysis,
which can be viewed as a specific form of abstract
interpretation~\cite{CousotC77}.  While the abstract interpretation
framework allows approximations to be performed in a more general way,
e.g., using relational abstract domains~\cite{Mine04} such as the
octagon~\cite{Mine01} and polyhedral~\cite{CousotH78} domains, so far,
it has not be adequately explored.  We plan to explore the use of
these abstract domains as part
of the future work.

Finally, the term \textit{differential verification} has been used
in the context of verifying a new version of a program with
respect to a previous version, which is treated as an ``oracle''~\cite{DAC13}.
In a sense, the truncated network is a ``new version'' of the original
network, and the original network can be thought of as an oracle.

\section{Conclusion}
\label{sec:conclusion}

We have presented a new method, named \diffNN{}, for differential
verification of two closely related neural networks.  It is capable of
formally proving the accuracy of a compressed network with respect to
the original network.  Internally, \diffNN{} relies on symbolic
interval analysis to more accurately compute and propagate differences
in the values of neurons of the two networks from the input to the
output, and then relies on the gradient difference to more accurately
compute the refinement.  Our experimental comparison of \diffNN{} with
state-of-the-art formal verification techniques shows that it can often
achieve two orders of magnitude speedup and produce many more proofs.

\section*{Acknowledgments}

This work was partially funded by the U.S. Office of Naval Research (ONR) under the grant N00014-17-1-2896.

\newpage\clearpage
\bibliography{diffNN}	

\appendix
\section{Optimization 3 Proofs}
Here we give formal proofs for the un-proven bounds in Section~\ref{sec:opt}.
As in section~\ref{sec:opt}, we rewrite Equations~\ref{eq:1} and~\ref{eq:2} as
\begin{align*}
\deq(n,d) & = \R{n + d} - \R{n}\\
\deq'(n', d) & = \R{n'} - \R{n' - d}.
\end{align*}
where $ n \in \ReluIN{\n}, n' \in \ReluIN{\np}, $ and $ d \in \ReluINdelta{\n} $.

\subsection{Upper Bound of First Case}
Recall this case is $ \LBconcrete{\ReluINdelta{\n}} \geq 0 $.
We derive the bound from Equation~\ref{eq:2} by dividing into two cases, and then combining their result
to get the lower bound.

\subsection*{Case 1: $ n' - d > 0 = n' > d $}
Based on the above constraint, we can simplify $ \deq'(n',d) $ to
\[
	\deq'(n',d) = \R{n'} - (n' - d).
\]
In addition, since we are considering the case where $ \LBconcrete{\ReluINdelta{\n}} \geq 0 $,
we have $ d \geq 0 $. Combining this with our case 1 constraint $ n' > d $,
we have
\[
	n' > d \wedge d \geq 0 \implies n' > 0.
\]
Thus, in case 1, we have that $ \deq'(n',d) $ is just
\[
	\deq'(n',d) = n' - (n' - d) = d.
\]

We also note that $ n' > 0 $ means that
\[
n' > d \iff max(0, n') > d
\]
because $ n' > 0 $ means that $ max(0, n') = n' $. We use this fact when we combine the two cases.

\subsection*{Case 2: $ n' - d \leq 0 = n' \leq d $}
The above constraint allows us to simplify $ \deq'(n',d) $ to
\[
	\deq'(n',d) = \R{n'} - 0 = \R{n'} = max(0, n').
\]
We also note that under our constraints we have
\[
n' \leq d \iff max(0, n') \leq d,
\]
which we will use when we combine the cases. To prove this, first observe that
\[
0 \leq d \wedge n' \leq d \implies max(0, n') \leq d.
\]
That is, if $ d $ is greater-than-or-equal to both 0 and $ n' $, then clearly
it is greater-than-or-equal to the max of the two. And for the other way,
\[
max(0, n') \leq d \implies  n' \leq d.
\]
That is, if $ d $ is greater-than-or-equal to the max of 0 and $ n' $, then clearly
it is greater-than-or-equal to $ n' $.

\subsection*{Combing the Two Cases}
Combining our two cases, we get the function
\[
	\deq'(n',d) = \begin{cases}
	d & n' > d\\
	max(0, n') & n' \leq d
	\end{cases}
\]
and then substituting with the equations we derived at the end of case 1 and 2, we get
\[
\deq'(n',d) = \begin{cases}
d & max(0, n') > d\\
max(0, n') & max(0, n') \leq d
\end{cases} = min(max(0, n'), d)
\]
Since the upper bound of both $ min() $ and $ max() $ occur when we take the upper bounds of their input variables, we get that the upper bound of $ \deq'(n',d) $ is
\begin{align*}
min(max(0, \UBconcrete{\ReluIN{\np}}), \UBconcrete{\ReluINdelta{\n}})\\
= min(\UBconcrete{\ReluIN{\np}}, \UBconcrete{\ReluINdelta{\n}}).
\end{align*}
We can remove the $ max() $ function because $ \np $ is non-linear, so
$ \UBconcrete{\ReluIN{\np}} > 0 $.

\subsection{Lower Bound of Second Case}
Recall this case is $ \UBconcrete{\ReluINdelta{\n}} \leq 0 $. We derive the lower bound from Equation~\ref{eq:1} by dividing into two cases. The proof is symmetric to the previous case.
\subsection*{Case 1: $ n + d > 0 = n > -d = d > -n $}
In this case, we can immediately simplify
\[
	\deq(n,d) = n + d - \R{n}.
\]
Then, since $ \UBconcrete{\ReluINdelta{\n}} \leq 0 $ implies $ d \leq 0 = -d \geq 0 $,
we can use our case 1 constraint $ n > -d $ to derive
\[
	n > -d \wedge -d \geq 0 \implies n > 0.
\]
Thus we can further simplify
\[
	\deq(n,d) = n + d - n = d.
\]

We also note that $ n > 0 = -n < 0 \implies min(0, -n) = -n $, so we also have
\[
	d > -n \iff d > min(0, -n).
\] 
We use this fact when combining the two cases.

\subsection*{Case 2: $ n + d \leq 0 = d \leq -n $}
In this case, we can immediately simplify
\[
	\deq(n,d) = 0 - \R{n} = -max(0, n) = min(0, -n).
\]
We also not here that
\[
	d \leq -n \wedge d \leq 0 \iff d \leq min(0, -n).
\]
We use this fact when combining the two cases.

\subsection*{Combining the Cases}
Combing our two cases, we get the function
\[
	\deq(n,d) = \begin{cases}
	d & d > -n\\
	min(0, -n) & d \leq -n
	\end{cases}
\]
and rewriting this equation using the inequalities we derived at the end of case 1 and 2
we get
\[
\deq(n,d) = \begin{cases}
d & d > min(0, -n)\\
min(0, -n) & d \leq min(0, -n)
\end{cases} = max(d, min(0, -n)).
\]
Since the lower bound of both $ min() $ and $ max() $ occur when we minimize their inputs, we get the lower bound of $ \deq(n,d) $ is
\begin{align*}
max(\LBconcrete{\ReluINdelta{\n}}, min(0, -\UBconcrete{\ReluIN{\n}}))\\
= max(\LBconcrete{\ReluINdelta{\n}}, -\UBconcrete{\ReluIN{\n}}).
\end{align*}
We can remove the $ min() $ function because $ \n $ is non-linear so $ -\UBconcrete{\ReluIN{\n}} < 0 $.

\subsection{Lower Bound of Third Case}
Recall this case is $ \LBconcrete{\ReluINdelta{\n}} < 0 < \UBconcrete{\ReluINdelta{\n}} $. We derive the lower bound from Equation~\ref{eq:1}. We divide into two cases and then combine them as done previously.

\subsection*{Case 1: $ n + d > 0 = d > -n $}
$ \deq(n,d) $ simplifies to
\[
	\deq(n,d) = n + d - \R{n}.
\]
We further divide into two sub-cases.

\subsection*{Case 1.1: $ n > 0 $}
$ \deq(n,d) $ further simplifies to
\[
	\deq(n,d) = n + d - n = d.
\]
Since we only care about the lower bound, observe that the lower bound of
case 1.1 occurs when we take the \textit{minimum} value of $ d $, which is \textit{always} less than 0.

\subsection*{Case 1.2: $ n \leq 0 $}
$ \deq(n,d) $ further simplifies to
\[
	\deq(n,d) = n + d - 0 = n + d.
\]
Observe that the lower bound cannot be less than 0 in case 1.2 because of the case 1 constraint.
This means the lower bound \textit{always} occurs in case 1.1, so we can safely ignore case 1.2. (But we emphasize this \textit{only} applies when evaluating $ \deq(n,d) $ for the lower bound).

\subsection*{Case 2: $ n + d \leq 0 = d \leq -n $}
$ \deq(n,d) $ further simplifies to
\[
	\deq(n,d) = 0 - \R{n} = -\R{n}.
\]
We consider the two cases of this function.
\subsection*{Case 2.1: $ n > 0 $}
$ \deq(n,d) $ becomes
\[
	\deq(n,d) = -n.
\]
The minimum value here occurs at the upper bound of $ n $, which is \textit{always} less than 0 because $ \n $ is non-linear.

\subsection*{Case 2.2: $ n \leq 0 $}
$ \deq(n,d) = 0 $ in this case. Since the lower bound of case 2.1 is \textit{always} less than 0, the minimum value will never occur in this case, so we can safely ignore it.

\subsection*{Combining the Cases}
We've shown that the minimum value occurs in either case 1.1 or case 2.1, which gives us the function (however only for the \textit{minimum} value and not the maximum)
\[
\deq(n,d) = \begin{cases}
d & d > -n\\
-n & d \leq -n
\end{cases} = max(-n,d).
\]
Evaluating this function for its lower bounds gives us 
\[
 max(-\UBconcrete{\ReluIN{\n}}, \LBconcrete{\ReluINdelta{\n}}).
\]

\subsection{Upper Bound of Third Case}
Recall this case is $ \LBconcrete{\ReluINdelta{\n}} < 0 < \UBconcrete{\ReluINdelta{\n}} $. We derive the upper bound from Equation~\ref{eq:2}. We divide into two cases and then combine them as done previously. This proof mirrors the proof for the lower bound of the third case.

\subsection*{Case 1: $ n' - d > 0 = n' > d $}
$ \deq'(n',d) $ simplifies to
\[
\deq'(n',d) = \R{n'} - n' + d.
\]
We further divide into two sub-cases.

\subsection*{Case 1.1: $ n' > 0 $}
$ \deq'(n',d) $ further simplifies to
\[
\deq'(n',d) = n' - n' + d = d.
\]
Since we only care about the upper bound, observe that the upper bound of
case 1.1 occurs when we take the \textit{maximum} value of $ d $, which is always greater than 0.

\subsection*{Case 1.2: $ n' \leq 0 $}
$ \deq'(n',d) $ further simplifies to
\[
\deq'(n',d) = 0 - n' + d = -n' + d.
\]
Our case 1 constraint $ n' - d > 0 = 0 > -n' + d $ implies the upper bound in case 1.2
can be no greater than 0. This means the upper bound $ \deq'(n',d) $ always occurs in case 1.1, so we
can ignore case 1.2.

\subsection*{Case 2: $ n' - d \leq 0 = n' \leq d $}
$ \deq'(n',d) $ further simplifies to
\[
\deq'(n',d) = \R{n'} - 0 = \R{n'}.
\]
We consider the two cases of this function.
\subsection*{Case 2.1: $ n' > 0 $}
$ \deq'(n',d) $ becomes
\[
	\deq'(n',d) = n',
\]
which has its maximum value at the upper bound of $ n' $.

\subsection*{Case 2.2: $ n' \leq 0 $}
$ \deq(n,d) = 0 $ in this case. Since the upper bound of $ n' $ is greater than 0, the maximum value will never occur in this case, so we can safely ignore it.

\subsection*{Combining the Cases}
We've shown that the maximum value occurs in either case 1.1 or case 2.1, which gives us the function (only for the \textit{maximum} value and not the minimum)
\[
\deq'(n',d) = \begin{cases}
d & n' > d\\
n' & n' \leq d
\end{cases} = min(n',d).
\]
Evaluating this function for its upper bound gives us 

\[
	min(\UBconcrete{\ReluIN{\np}}, \UBconcrete{\ReluINdelta{\n}}).
\]

\end{document}